\documentclass{article}

\usepackage[table]{xcolor}
\usepackage[final]{corl_2026}
\usepackage{amsmath,amssymb}
\usepackage{array}
\usepackage{booktabs}
\usepackage{colortbl}
\usepackage{enumitem}
\usepackage{float}
\usepackage{graphicx}
\usepackage{tabularx}
\usepackage{makecell}
\usepackage{multirow}
\usepackage{tikz}
\usepackage{wrapfig}
\usepackage{algorithm}
\usepackage{algpseudocode}
\usetikzlibrary{arrows.meta,positioning,shapes.geometric}

\title{Graph-Reweighted Behavior Cloning with Human Interventions for Precise Robotic Manipulation}
\author{
Xiaoman Li$^{1,2}$
~~
Zhiqiang Yang$^{2,3}$
~~
Jiayu Hu$^{2,4}$
~~
Jipeng Kong$^{2,5}$ \\
\textbf{Yang Zhang}$^{2,6}$
~~
\textbf{Jingwei Zhang}$^{2,3}$
~~
\textbf{Fangzheng Yan}$^{2}$
~~
\textbf{Chunhua Shen}$^{1\dagger}$
~~
\textbf{Chenjia Bai}$^{2,7\dagger}$
\\[0.253cm]
$^1$ Zhejiang University, State Key Lab of CAD \& CG \\
$^2$ The Institute of Artificial Intelligence, China Telecom (TeleAI) \\
$^3$ South China University of Technology
~~~
$^4$ University of Science and Technology of China \\
$^5$ ShanghaiTech University
~~~
$^6$ Tsinghua University
~~~
$^7$ Gamma Robotics ($\gamma$) \\
$^{\dagger}$Corresponding Authors
}

\newcommand{\D}{\mathcal{D}}
\newcommand{\Ddemo}{\mathcal{D}_{\mathrm{demo}}}

\newcommand{\Drollout}{\mathcal{D}_{\mathrm{rollout}}}
\newcommand{\Dr}{\mathcal{D}_r}

\newcommand{\Gv}{G_{\mathrm{V}}}

\newcommand{\Rrec}{R_{\mathrm{rec}}}

\newcommand{\ind}{\mathbb{I}}

\hypersetup{hypertexnames=false}

\begin{document}
\maketitle

\vspace{-6pt}
\begin{abstract}
Long-horizon precise manipulation produces mixed-quality data for post-training vision-language-action (VLA) models: even successful demonstrations and intervention-assisted rollouts may contain detours, hesitations, failed attempts, and imperfect recoveries.
The key challenge is to identify which action chunks actually advance the task, rather than uniformly cloning or discarding imperfect trajectories.
We propose Graph-Reweighted Behavior Cloning (GR-BC), an offline iterative post-training framework that builds a task-conditioned goal-reachability graph over demonstrations and iteratively collected rollouts, including both successful and failed episodes with human interventions.
Reverse graph search computes minimal remaining steps to task success within the dataset support, and each action chunk is weighted by how much its execution reduces this potential.
This consequence-based weighting reuses recoverable segments from failed trajectories, leverages interventions as recovery paths, and down-weights locally regressive chunks even inside successful trajectories.
New rollouts are incorporated by appending them to the graph and rerunning search, without training a reward model, action-conditioned critic, or progress predictor.
Across four precise manipulation tasks and three VLA backbones, GR-BC improves single-round supervised fine-tuning and reaches 91\% success after a few offline iterations, without trajectory comparisons or progress labels. More details and videos are available at \url{https://gr-bc.github.io/}
\end{abstract}


\vspace{-3pt}
\section{Introduction}
\vspace{-3pt}

Post-training vision-language-action (VLA) models~\citep{kim2024openvla,zitkovich2023rt,black2024pi_0,black2025pi_,bjorck2025gr00t,zhai2025igniting,wu2026pragmatic,zhang2026prts,yang2025steering,huang2025notvla,liu2026stamo} for long-horizon precise manipulation rarely start with ideally clean expert datasets. Demonstrations collected by human experts can contain suboptimal or undesired behaviors~\citep{yin2024offline,mao2026arm,liu2026perfect}, such as hesitations, overshoots, alignment detours, or recoveries from an earlier failed attempt.
Intervention-assisted rollouts~\citep{kelly2019hg} are even more heterogeneous, mixing autonomous failures with delayed expert corrections and recoveries.
Conversely, failure rollouts are not uniformly useless as they can also traverse goal-reachable states or contain short trajectory segments that steer the system back toward a viable path.
Thus, the core challenge is essentially chunk-level credit assignment: whether an action chunk is worth imitating depends on how much task progress it induces when executed from the current state, rather than the final outcome of the entire trajectory.
This motivates our central question: how can we effectively extract these progress-making action chunks from mixed-quality data, especially across iterative human-in-the-loop deployment rounds, to enhance sample efficiency for VLA post-training?

\begin{figure}[!t]
\centering
\makebox[\linewidth][c]{\includegraphics[width=1\linewidth]{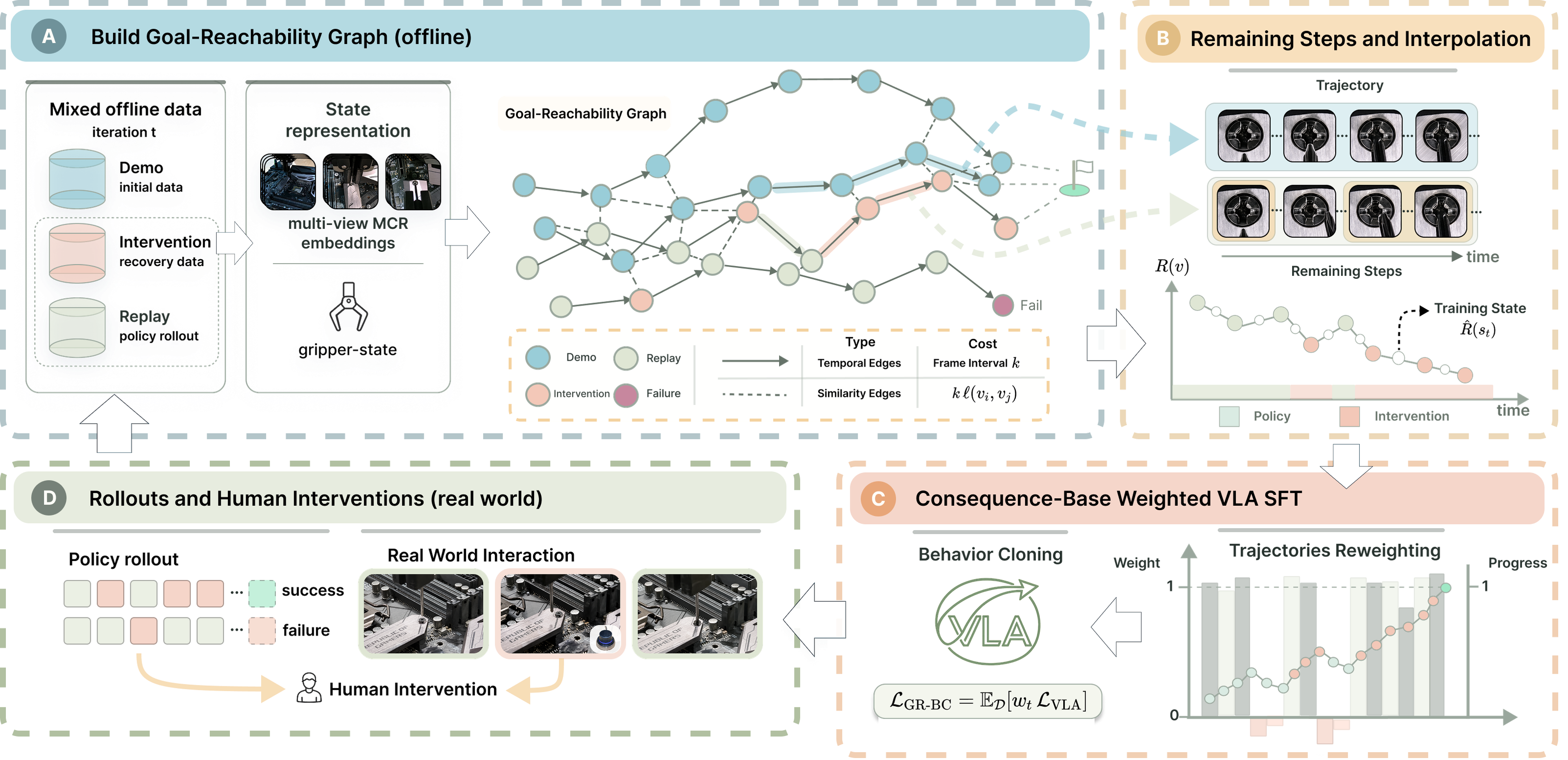}}
\vspace{-1.2em}
\caption{Overview of the GR-BC framework. (A) We construct a task-conditioned goal-reachability graph from mixed-quality offline data. (B) Reverse graph search estimates minimal remaining steps to task success, providing dense potential signals. (C) Action chunks are weighted by these consequence-based potentials for VLA SFT. (D) New real-world rollouts, including failed episodes and human interventions, are iteratively incorporated to refine the graph and policy.}
\label{fig:pipeline}
\vspace{-0.38em}
\end{figure}

Existing post-training and progress estimation approaches struggle to provide this fine-grained signal within the mixed-quality data.
Standard supervised fine-tuning (SFT) naively treats every action chunk as a positive target, thereby inadvertently cloning undesired behaviors from expert demonstrations or intervention-assisted rollouts~\citep{yin2024offline,liu2025robot}. Many progress estimation methods implicitly assume that task progress evolves monotonically along successful trajectories~\citep{chen2025sarm}, which hardly holds when trajectories contain correction behaviors or pauses~\citep{yin2024offline,mao2026arm}.
Although recent reward modeling approaches attempt address this by predicting chunk-level progress, they are inevitably bottlenecked by two limitations: (i) an over-reliance on heavy human or VLM annotations, such as dense progress labels, subtask boundaries, and trajectory comparisons~\citep{chen2025sarm,mao2026arm,feng2026procvlm,liang2026robometer}; and (ii) the out-of-distribution (OOD) generalization gap inherent in large-scale pre-trained reward models, which often fail to provide reliable zero-shot progress signals when confronted with the precise visual nuances of unseen downstream tasks~\citep{lee2026roboreward,tan2025robo,chen2026topreward,wu2026large}.
While graph-search imitation~\citep{yin2024offline} avoids learning a neural progress predictor, it uses graph search primarily to retrieve high-value neighboring actions for imitation, rather than to quantify how much the action chunk actually executed at a state moves the system toward success.

Our core insight is that the quality of an action chunk should be measured by its consequence: specifically, whether its execution reduces the minimal remaining steps to task completion.
Crucially, this signal naturally penalizes local regressions inside successful trajectories, while simultaneously up-weighting valid progress-making chunks hidden within an overall failed rollouts.

To this end, we propose GR-BC, a graph-based framework for iterative offline VLA post-training over long-horizon precise manipulation tasks. GR-BC constructs a task-conditioned goal-reachability graph over demonstrations and intervention-assisted rollouts using multi-view MCR~\citep{jiang2025robots} embeddings to retrieve candidate visual neighbors and conservative similarity edges to connect dynamically compatible states. Instead of ranking neighbors or training a progress predictor, GR-BC runs reverse graph search to estimate minimal remaining steps and interpolates this potential to dense training states. The policy is then optimized only on physically executed action chunks, with each chunk's contribution scaled by a precomputed weight proportional to its progress improvement. Because the graph is non-parametric, each offline iteration only appends new rollouts and reruns search, with no reward model, critic, progress predictor, architecture change, or supervised-loss modification.

The contributions are:
\vspace{-0.7em}
\begin{itemize}
    \item \textbf{Graph-Retrieved Progress (GRP):}
    We propose a non-parametric progress estimation method that constructs a task-conditioned reachability graph and retrieves minimal remaining steps to success, avoiding reward models, critics, progress labels, or VLM annotations.

    \item \textbf{Graph-Reweighted Behavior Cloning (GR-BC):}
    We introduce a graph-reweighted behavior cloning objective that preserves the standard supervised VLA loss while reweighting each executed action chunk using GRP. This consequence-based reweighting reuses recovery chunks from failed rollouts while suppressing regressive chunks in successful trajectories.

    \item \textbf{Iterative Offline Graph-Guided VLA Adaptation:}
    We validate GR-BC in human-in-the-loop offline iterative training by updating the reachability graph with intervention-assisted rollouts and reusing them in subsequent training rounds, achieving over 90\% success on multiple long-horizon precise manipulation tasks with multiple VLA backbones.
\end{itemize}
\vspace{-0.28em}
\vspace{-6pt}
\section{Related Work}
\label{sec:related_work}
\vspace{-6pt}

\paragraph{Progress Rewards for Robotic Manipulation.}
Long-horizon precise manipulation often relies on dense progress signals to provide supervision beyond sparse task success. 
Some methods infer progress from temporal order, goal proximity, or video ranking~\citep{wu2021learning,lee2021generalizable,yang2024rank2reward}, while others learn stage- or subtask-aware progress models from segmented demonstrations or language annotations~\citep{chen2025sarm,mu2024drs,kim2025subtask}.
More recent work uses foundation models or process reward models to estimate visual task progress from language, videos, or VLM judgments~\citep{lee2026roboreward,tan2025robo,hung2025victor,ma2023liv,alakuijala2024video,ma2025vision,zhang2025rewind}. ARM~\citep{mao2026arm} is closest to our motivation, replacing absolute progress scores with relative advantage labels for progressive, regressive, and stagnant behavior.
However, these approaches typically require learned reward models, progress labels, stage annotations, VLM scoring, or temporal-order assumptions. GRP instead retrieves progress directly from graph reachability and reverse search, without learning a reward or progress function.
\vspace{-0.2em}
\paragraph{Weighted Behavior Cloning for VLA Post-training.}
Many RL-based VLA post-training methods improve downstream task performance while preserving the generalization inherited from pretrained VLA backbones through constrained updates~\citep{wagenmaker2025steering,xiao2025self,li2025gr,chen2025conrft,wang2026learning,liu2026perturbot}. Within this broader paradigm, weighted behavior cloning improves policy learning from heterogeneous data by assigning higher likelihood weights to actions estimated to be more valuable. Many weighted imitation and offline RL methods derive training weights from estimated values or advantages~\citep{peng2019advantage,nair2020awac,kostrikov2021offline,chen2020bail}. Recent robot and VLA post-training methods similarly reweight behavior cloning using reward, advantage, intervention, or off-policy value estimates~\citep{chen2025sarm,mao2026arm,liu2025robot,yang2026aloe,intelligence2025pi,su2026ig}.
Unlike prior weighted BC methods, GR-BC does not estimate action values, rewards, or advantages.
Instead, it uses a non-parametric, graph-derived progress potential to derive weights from the observed reduction in minimal remaining steps, directly measuring the consequence of executed action chunks.

\begin{figure}[!t]
\centering
\makebox[\linewidth][c]{\includegraphics[width=1\linewidth]{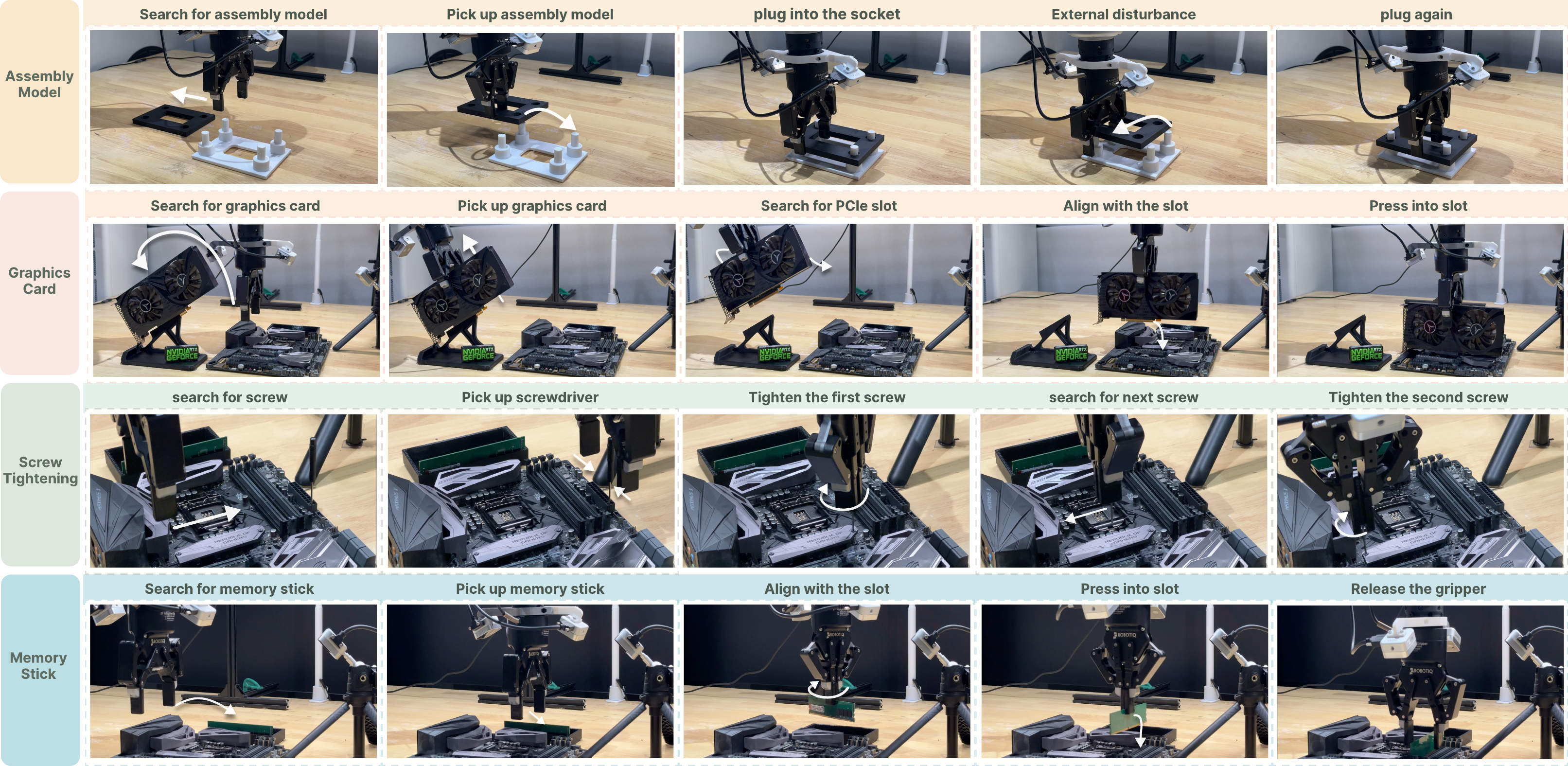}}
\caption{Real-world evaluation tasks.
We evaluate GR-BC on four long-horizon precise manipulation tasks requiring precise visual alignment, contact-rich interactions, and recovery from execution errors.}
\label{fig:task_overview}
\end{figure}

\vspace{-6pt}
\section{Preliminaries}
\vspace{-6pt}

\paragraph{Problem Formulation.} We study interactive imitation learning as an iterative offline VLA post-training procedure. Given an iteration round $r$, the mixed-quality data pool $\mathcal{D}^r$ is defined as:
\begin{equation}
    \mathcal{D}^r = \mathcal{D}_{\mathrm{demo}} \cup \mathcal{D}_{\mathrm{rollout}}^{1:r}, \qquad \mathcal{D}_{\mathrm{rollout}}^{1:r} = \mathcal{D}_{\mathrm{success\text{-}rollout}}^{1:r} \cup \mathcal{D}_{\mathrm{fail\text{-}rollout}}^{1:r}.
\end{equation}
Here, rollouts are labeled by their final task outcomes and may contain human interventions at arbitrary moments. Each trajectory $\tau = (s_0,a_0,\dots,s_n,a_n) \in \mathcal{D}^r$ is annotated with metadata including a language instruction $g$, the overall outcome, and intervention timestamps. The state $s_t$ aggregates three RGB views $(o_t^1,o_t^2,o_t^3)$, proprioception, and a binary gripper state $q_t^{\mathrm{grip}}$. While $s_t$ forms the basis for graph node construction, the VLA policy $\pi_\phi$ is optimized to predict future action chunks $a_{t:t+N}$ conditioned on $(s_t,g)$. As trajectory-level outcomes do not guarantee localized action optimality, our goal is to assign a fine-grained quality weight to each observed chunk.

\paragraph{Offline Weighted VLA Post-training.}
\label{sec:weighted_vla}
We use weighted supervised post-training over real-observed chunks:
\begin{equation}
    \mathcal L(\phi)
    =
    \mathbb{E}_{(s_t,g,a_{t:t+N-1})\sim\Dr}
    \left[
        w_t\,
        \mathcal{L}_{\mathrm{VLA}}
        \left(
            \pi_\phi,
            s_t,
            g,
            a_{t:t+N-1}
        \right)
    \right],
\end{equation}
where $w_t\ge 0$ is a sample-level weight (recovering standard SFT when $w_t=1$), and $\mathcal{L}_{\mathrm{VLA}}$ denotes a standard imitation objective, such as behavioral cloning, denoising, or flow-matching~\citep{chi2025diffusion}.
GR-BC adopts this interface by deriving $w_t$ from GRP; thus, the graph evaluates topological state potentials instead of actions~(Fig.~\ref{fig:pipeline}), leaving VLA training strictly supervised by the original dataset chunks.

\vspace{-6pt}
\section{Methodology}
\vspace{-6pt}

As illustrated in Fig.~\ref{fig:pipeline}, GR-BC first builds a task-conditioned goal-reachability graph over mixed-quality trajectories (Sec.~\ref{sec:value_graph}), retrieves graph-retrieved progress from reverse-search remaining-step potentials (Sec.~\ref{sec:progress_estimation}), and uses the resulting progress improvement to reweight action chunks for iterative VLA post-training (Sec.~\ref{sec:weighted_BC}).

\vspace{-6pt}
\subsection{Goal-reachability Graph Construction}\label{sec:value_graph}
\vspace{-6pt}

To leverage recoverable segments within mixed-quality historical rollouts, we structure the cumulative offline dataset $\mathcal{D}^r$ into a task-conditioned goal-reachability graph $\mathcal{G} = (\mathcal{V}, \mathcal{E})$. 

\begin{wrapfigure}{r}{0.45\linewidth}
\vspace{-0.6em}
\centering
\includegraphics[width=0.95\linewidth]{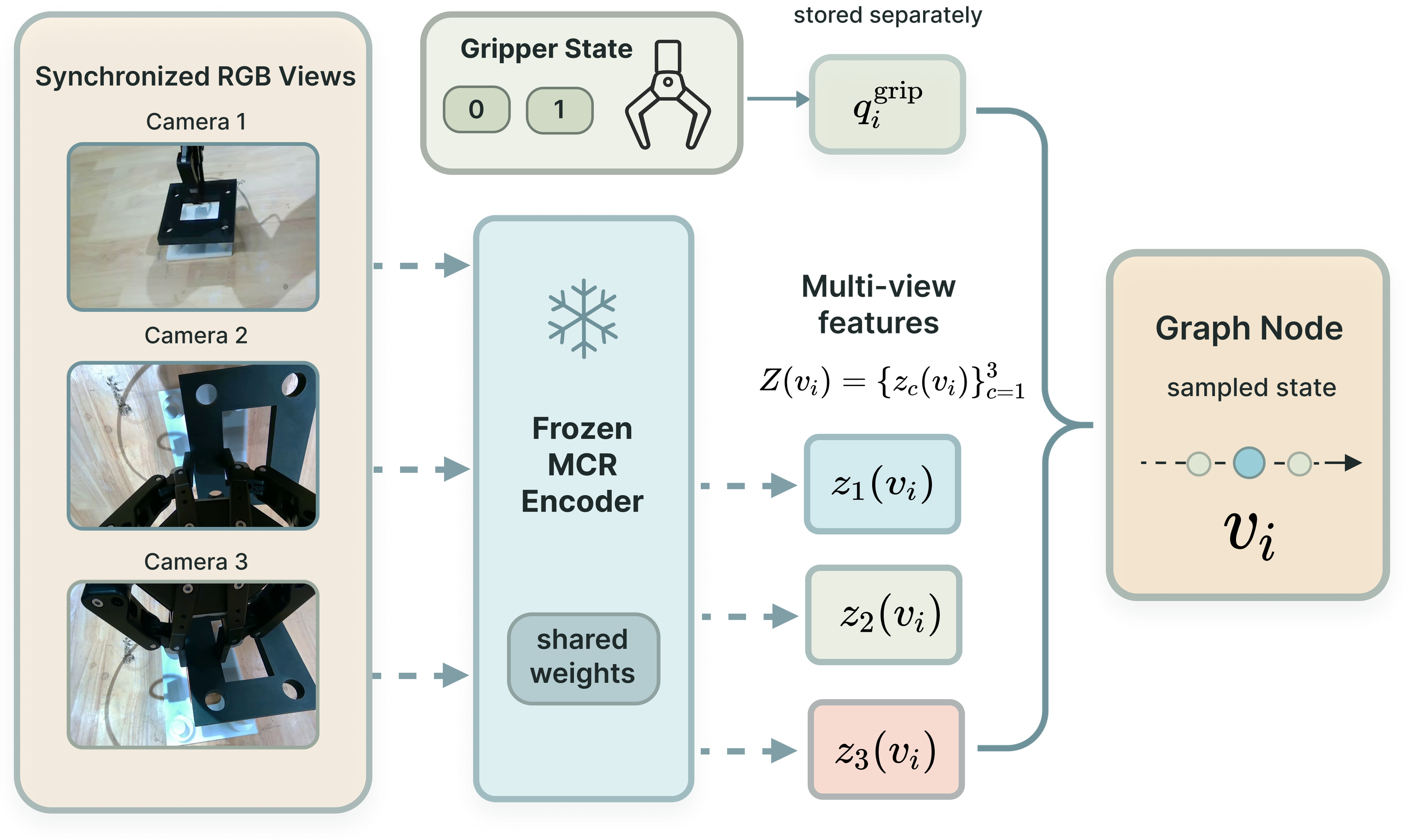}
\vspace{-0.6em}
\caption{Multi-view node encoding for graph construction.}
\label{fig:encoder}
\vspace{-1em}
\end{wrapfigure}

\textbf{Node set $\mathcal{V}$:} To maintain graph sparsity, trajectories are uniformly subsampled every $k$ frames while strictly retaining all terminal and successful states: $\mathcal{V} = \{s_t \mid t = 0, k, 2k, \dots\} \cup \mathcal{V}_{\mathrm{terminal}} \cup \mathcal{V}_{\mathrm{success}}$. Each node $v_i \in \mathcal{V}$ aggregates synchronized multi-view visual embeddings $Z(v_i) = \{z_c(v_i)\}_{c=1}^3$, where $z_c(v_i) = f_{\mathrm{MCR}}(o_c(v_i))$ via a frozen manipulation-centric encoder~\citep{jiang2025robots}, as shown in Fig.~\ref{fig:encoder}.

Crucially, the binary gripper state $q_i^{\mathrm{grip}}$ is maintained separately from $Z(v_i)$ to act as a hard structural filter during subsequent edge construction, preventing the soft alignment of dynamically incompatible states.

\textbf{Temporal edges $\mathcal{E}$.}
To encode physically executed transitions, directed edges are established between consecutive intra-trajectory nodes as
$v_i \to v_{i+k}$ with cost $c(v_i, v_{i+k}) = k$, ensuring that the cumulative path cost naturally quantifies the elapsed physical frames along the observed historical rollouts.

\textbf{Cross-trajectory similarity edges $\mathcal{E}_{\mathrm{sim}}$.} To bridge independent trajectories and establish goal-reachability for failed or intervened rollouts, we introduce undirected similarity edges $\mathcal{E}_{\mathrm{sim}}$ governed by multi-view consensus, gripper-state consistency, and local distance adaptability. We first construct a candidate pair set $C_{\mathrm{mv}}$ by intersecting the mutual $K$-nearest neighbors ($K$-NN) queried within each camera view's MCR feature space, effectively eliminating spurious single-view matches. For each candidate pair, the raw feature distance is normalized by a local resolution metric to ensure a consistent physical interpretation across the embedding space.

For camera $c$, let $\mathcal{N}(v)$ denote the immediate temporal neighbors of node $v$. We define the local scale as:
\begin{equation}
\sigma_c(v) = \frac{1}{|\mathcal N(v)|} \sum_{q \in \mathcal N(v)} \|z_c(v) - z_c(q)\|_2,
\end{equation}
which captures how visually distant a single transition around $v$ on the graph is.

\begin{wrapfigure}{l}{0.45\linewidth}
    \centering
    \includegraphics[width=0.8\linewidth]{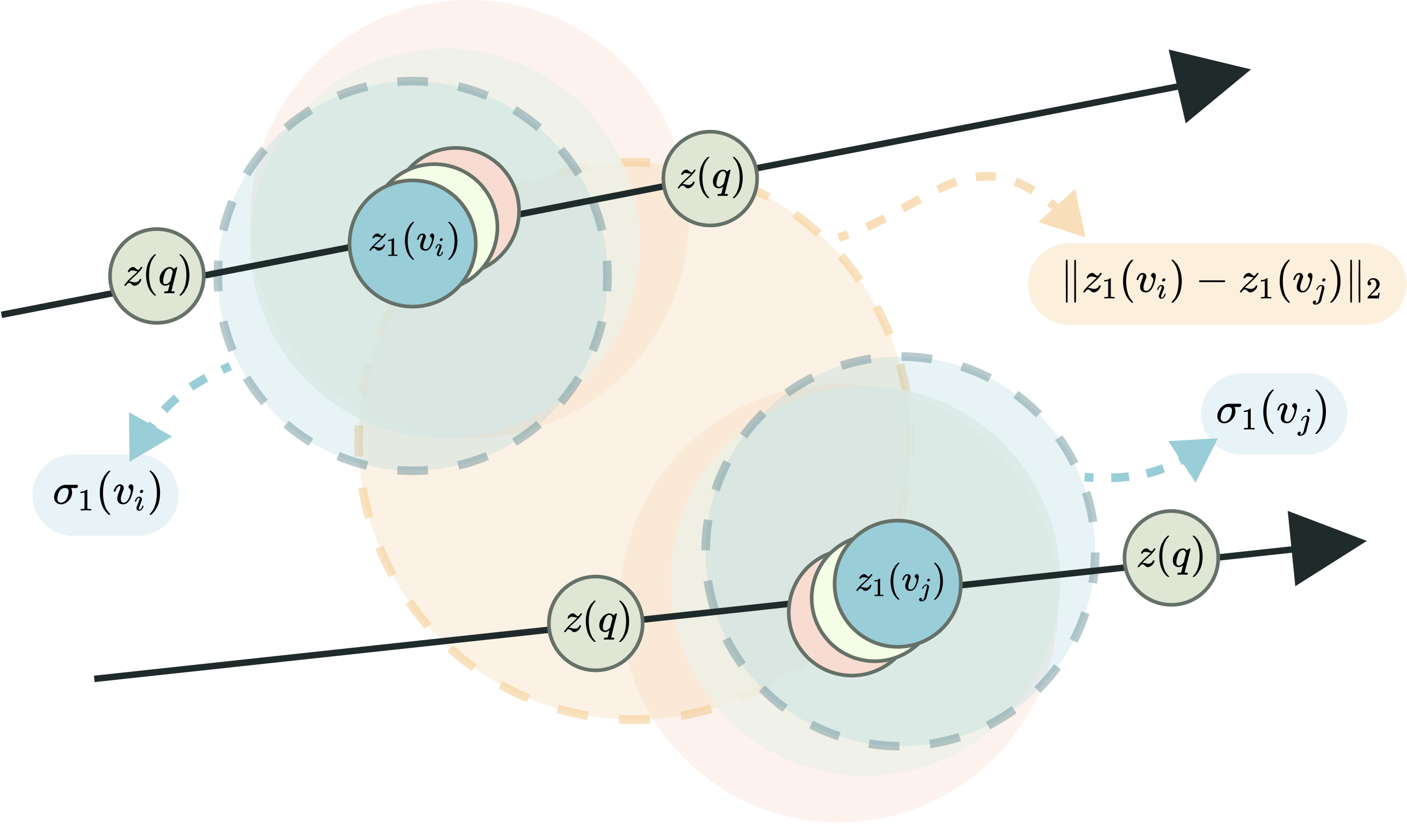}
    \caption{\textbf{Cross-trajectory similarity computation.} We compute the pair-wise feature distance between states $v_i$ and $v_j$, normalized by their local adaptive scales $\sigma_c(v)$, to establish connectivity that reflects consistent physical state compatibility.}
    \label{fig:similarity}
    \vspace{-1.65em}
\end{wrapfigure}

As illustrated in Fig.~\ref{fig:similarity}, the multi-view adaptive edge length is then normalized by the maximum relative distance across all camera views:
\begin{equation}
\ell(v_i, v_j) = \max_{c \in \{1, 2, 3\}} \frac{\|z_c(v_i) - z_c(v_j)\|_2}{\epsilon_c + (\sigma_c(v_i) + \sigma_c(v_j))/2},
\end{equation}
where $\epsilon_c$ is a small constant for numerical stability.

Combining these three criteria, the similarity edge set is formulated as
\begin{equation}
\mathcal{E}_{\mathrm{sim}} = \{(v_i, v_j) \in C_{\mathrm{mv}} \mid q_i^{\mathrm{grip}} = q_j^{\mathrm{grip}} \;\land\; \ell(v_i, v_j) \le \alpha_{\mathrm{edge}}\},
\end{equation}

with cost proportional to its adaptive length $c(v_i, v_j) =k \cdot \ell(v_i, v_j)$, where $\alpha_{\mathrm{edge}}$ is a pre-defined threshold. Since temporal edges have cost $k$, this normalized similarity cost preserves the relative penalty of cross-trajectory bridges while still allowing failure and intervention states to reconnect to success paths when only similarity evidence suffices.

\vspace{-6pt}
\subsection{Remaining Steps and Interpolation}\label{sec:progress_estimation}
\vspace{-6pt}

Given $\Gv$, we compute a goal-reachable remaining-step estimate $\Rrec(v)$, which is subsequently dense-interpolated across all trajectory frames to support chunk-level post-training.

\textbf{Reverse Path Search:}
To reference graph distance to task completion, we introduce a virtual target sink $g_\star$ connected from all successful terminal nodes via zero-cost directed edges $v_{\mathrm{success}} \rightarrow g_\star$. Conversely, failed or incomplete terminal nodes are excluded from $g_\star$, forcing unsuccessful trajectories to route through valid similarity bridges to establish goal-reachability. Executing Dijkstra's algorithm from $g_\star$ on the reversed graph $\Gv$ yields the minimum task-completion cost for any node $v$: $\Rrec(v) = \min_{\pi: v \to g_\star} \sum_{e \in \pi} c(e)$. Nodes with no path to $g_\star$ are assigned $\Rrec(v) = \infty$, resulting in a zero training weight. This constraint enforces a conservative data-reuse policy, filtering out ungrounded, suboptimal behaviors while salvaging locally successful segments.

\textbf{Dense Interpolation:} To bridge the gap between sparse graph nodes and the dense temporal supervision required by VLA training, any state $s_t$ situated between two adjacent trajectory nodes $v_i$ and $v_{i+k}$ (where $i \le t \le i+k$) is evaluated via linear interpolation:
\begin{equation}
\lambda_t = \frac{t - i}{k}, \qquad \hat{R}_{\mathrm{rec}}(s_t) = (1 - \lambda_t) \Rrec(v_i) + \lambda_t \Rrec(v_{i+k}).
\end{equation}
This procedure identically extends to any future horizon state $s_{t+N}$, ensuring dense alignment for the chunk-level loss function detailed in Sec.~\ref{sec:weighted_BC}.

\vspace{-6pt}
\subsection{Weighted SFT and Iterative Offline Update}\label{sec:weighted_BC}
\vspace{-6pt}

Leveraging the dense remaining-step estimates, we formalize a chunk-level reweighting mechanism and establish the weighted SFT objective for iterative offline policy optimization.

\begin{figure}[t]
\centering
\includegraphics[width=\linewidth]{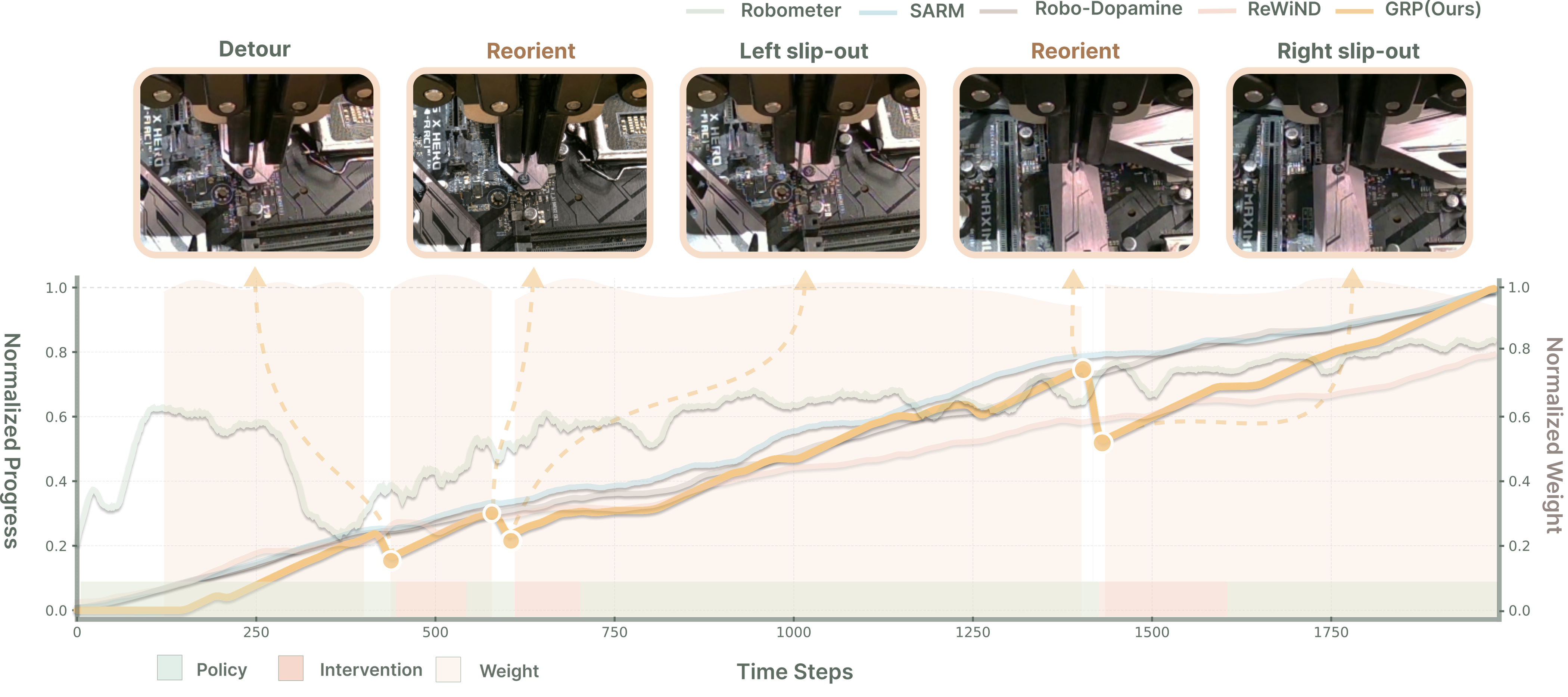}
\vspace{-1.2em}
\caption{\textbf{Comparison of progress estimation and transition weights.} GRP (ours) provides more consistent progress signals than reward modeling baselines (SARM, Robometer, Robo-Dopamine, ReWiND). Our consequence-based weights effectively isolate task-advancing segments from local regressions or failed behaviors.}
\label{fig:progress_model_comparison}
\end{figure}

\textbf{Chunk Consequence Evaluation:} To eliminate task-dependent scale variances across trajectories of varying lengths, the dense potential $\hat{R}_{\mathrm{rec}}(s)$ is first normalized into a bounded progress score $p(s) \in [0, 1]$:
\begin{equation}
    p(s) = \operatorname{clip}\left(1 - \frac{\hat{R}_{\mathrm{rec}}(s)}{R_{\max}}, \, 0, \, 1\right),
\end{equation}
where $R_{\max}$ denotes an upper percentile of finite $\Rrec$ values across the data pool to guard against extreme outliers. For an action chunk $a_{t:t+N-1}$ executed at state $s_t$, its trajectory impact over a lookahead horizon $N$ is then evaluated via two complementary metrics: $\Delta_t = p(s_{t+N}) - p(s_t), \quad d_t = p(s_t) - \min_{\tau \in [t, \, t+N]} p(s_\tau)$, where the net progress $\Delta_t$ measures endpoint progress, and the regression depth $d_t$ exposes transient intra-chunk violations. Explicitly tracking $d_t$ prevents the policy from cloning unstable, highly sub-optimal behaviors that superficially terminate with a positive net progress $\Delta_t$.

\textbf{Consequence-Modulated Optimization:} Chunks exhibiting stable forward progress are selectively prioritized via a step-wise transition weight $w_t$:
\begin{equation}
w_t =
\begin{cases}
\operatorname{clip}\left(1 + \beta_w \tanh(\gamma \Delta_t), \, w_{\min}, \, w_{\max}\right), & d_t \le \eta \land \Delta_t > \delta_{\min}, \\[4pt]
0, & \text{otherwise,}
\end{cases}
\end{equation}
where $\beta_w$ and $\gamma$ calibrate the progress bonus amplitude and sensitivity, bounded by $[w_{\min}, w_{\max}]$. Crucially, the zero-weight branch masks suboptimal gradients while retaining the corresponding states for graph connectivity.
GR-BC then applies the weighted VLA objective in Sec.~\ref{sec:weighted_vla} using consequence-derived weights, with the compact form \(\mathcal{L}_{\mathrm{GR\text{-}BC}}=\mathbb{E}_{\mathcal{D}}\!\left[w_t\,\mathcal{L}_{\mathrm{VLA}}\right]\), where \(w_t\) acts as a sample-level gradient multiplier without modifying the VLA architecture.

\textbf{Iterative Offline Update:} At round $r$, new rollouts $\mathcal{D}_{\mathrm{rollout}}^r$ are integrated into the data pool. GR-BC then increments the graph with the new rollouts and re-executes the reverse Dijkstra search to refresh $w_t$ for round $r+1$. Crucially, this pipeline shifts the computational bottleneck from costly neural network retraining to efficient graph lookups.

\vspace{-6pt}
\section{Experiments}
\label{sec:experiments}
\vspace{-6pt}
In this section, we answer three questions: %
\vspace{-0.3em}
\begin{itemize}[leftmargin=1.5em, labelsep=0.5em]
    \item \textbf{Q1:} Does GRP provide a more accurate progress signal for precise manipulation?
    \vspace{-0.1em}
    \item \textbf{Q2:} Does GR-BC improve VLA post-training on mixed-quality human-in-the-loop rollouts?
    \vspace{-0.1em}
    \item \textbf{Q3:} How do different graph-construction choices influence the quality of graph-retrieved progress?
\end{itemize}

\vspace{-0.3em}
We evaluate GR-BC on four real-world manipulation tasks shown in Fig.~\ref{fig:task_overview}. Additional implementation details are provided in Appendix~\ref{app:A}.

\begin{table}[H]
\centering
\caption{Progress-model interfaces and evaluation results.}
\label{tab:progress_method_interfaces}
\scriptsize
\setlength{\tabcolsep}{3pt}
\renewcommand{\arraystretch}{0.88}
\resizebox{\linewidth}{!}{
\begin{tabular*}{\linewidth}{@{\extracolsep{\fill}}lccccccc}
\toprule
Method & $\Delta_{\mathrm{SF}}\uparrow$ & $\rho_{\mathrm{dev}}\uparrow$
& Class. SE$\uparrow$ & Class. FE$\uparrow$ & Multi-cam. & Failure data & Progress labels \\
\midrule
ReWiND~\citep{zhang2025rewind} & 0.003 & 0 & 30/30 & 1/30 & -- & -- & -- \\
SARM~\citep{chen2025sarm} & 0.24 & 0.48 & 26/30 & 14/30 & -- & -- & \checkmark \\
Robo-Dopamine~\citep{tan2025robo} & 0.29 & 0.59 & 29/30 & 8/30 & \checkmark & -- & \checkmark \\
RoboMeter~\citep{liang2026robometer} & 0.159 & 0.69 & 30/30 & 6/30 & -- & \checkmark & \checkmark \\
\rowcolor[gray]{0.82}
GRP (ours) & 0.97 & 0.92 & 30/30 & 30/30 & \checkmark & \checkmark & -- \\
\bottomrule
\end{tabular*}}
\end{table}

\begin{figure}[H]
\centering
\includegraphics[width=\linewidth]{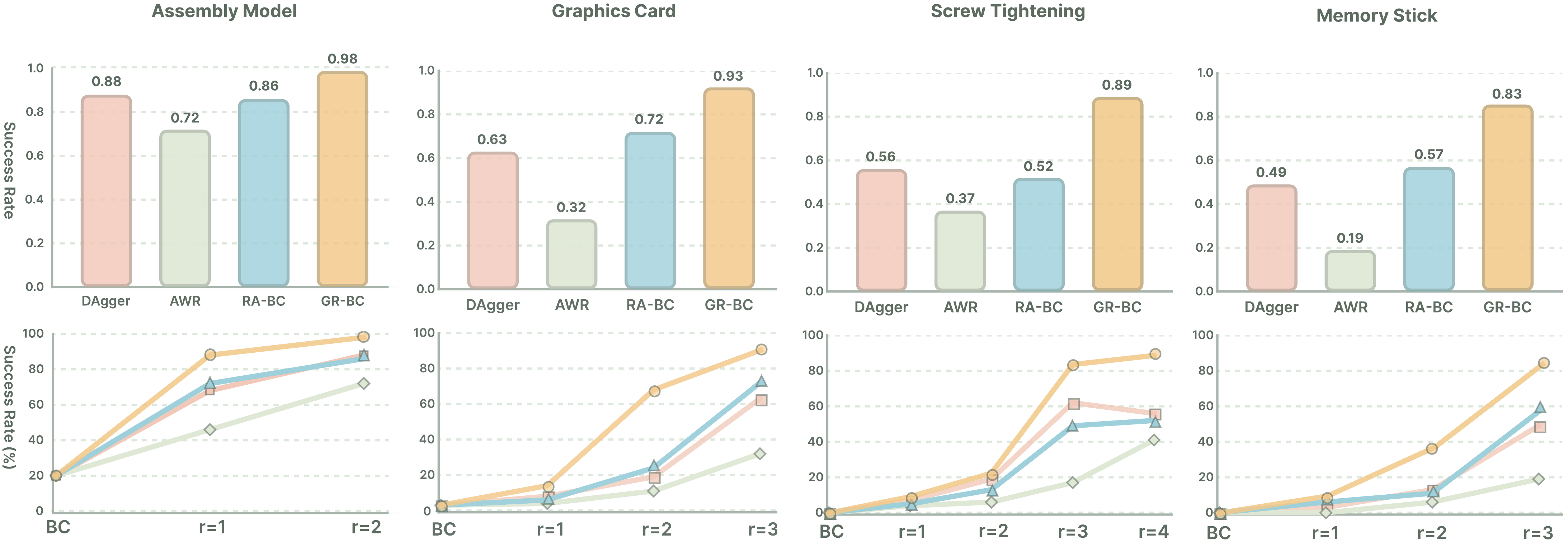}
\vspace{-1em}
\caption{Policy improvement across offline post-training rounds. GR-BC achieves more consistent gains than DAgger, AWR, and RA-BC on precise assembly tasks.}
\label{fig:policy_learning_results}
\end{figure}

\vspace{-6pt}
\subsection{Progress Estimation Evaluation}
\vspace{-6pt}

As visualized in Fig.~\ref{fig:progress_model_comparison}, we evaluate GRP against representative progress estimation baselines, including ReWiND~\citep{zhang2025rewind}, SARM~\citep{chen2025sarm}, Robo-Dopamine~\citep{tan2025robo}, and RoboMeter~\citep{liang2026robometer}, on mixed-quality trajectories collected across policy-training iterations. Table~\ref{tab:progress_method_interfaces} summarizes their data interfaces and evaluation results. Additional details of the progress-function evaluation protocol are provided in Appendix~\ref{app:B}.

\paragraph{Evaluation protocol.}
We evaluate each method under its intended supervision setting rather than enforcing a unified training protocol. Training and validation data are constructed according to each method's original assumptions, while evaluation trajectories remain fixed.
To avoid injecting additional subjective labels, we do not annotate any partial-success states or per-frame progress labels.
Method-specific assumptions are summarized in Table~\ref{tab:progress_method_interfaces}.
For each task, we evaluate 30 successful (SE) and 30 failed (FE) episodes. Following the SARM classification protocol~\citep{chen2025sarm}, we report trajectory-level classification results over successful and failed episodes. 
We also report the success--failure progress gap \(\Delta_{\mathrm{SF}}\), computed as the difference between final normalized progress on SE and FE episodes.
Finally, we select 100 manually localized deviation windows from successful trajectories. These windows capture locally regressive behaviors, including teleoperation mistakes and pre-intervention deviations. We define $\rho_{\mathrm{dev}}$ as the fraction of deviation windows whose start-to-end progress drop exceeds a threshold:
\begin{equation}
    \rho_{\mathrm{dev}}
    =
    \frac{1}{|\mathcal{W}_{\mathrm{dev}}|}
    \sum_{w=(t_s,t_e)\in\mathcal{W}_{\mathrm{dev}}}
    \mathbb{I}\left[
    p(s_{t_s}) - p(s_{t_e}) > \delta_{\mathrm{dev}}
    \right],
\end{equation}

where $t_s$ and $t_e$ denote the manually localized start and failure frame of window $w$, and $\delta_{\mathrm{dev}}$ is the minimum progress drop required for detection.
Higher $\rho_{\mathrm{dev}}$ indicates better sensitivity to locally regressive manipulation behavior.

\vspace{-6pt}
\subsection{Policy Improvement Evaluation}
\vspace{-6pt}
We use X-VLA~\citep{zheng2025x} as our primary experimental backbone, as its efficiency enables multiple offline post-training rounds. We also evaluate GR-BC with \(\pi_{0.5}\)~\citep{black2025pi_} and SmolVLA~\citep{shukor2025smolvla}, with these additional experiments detailed in Appendix~\ref{app:C}. All methods are warm-started from the same BC checkpoint trained on successful demonstrations. We compare GR-BC with three baselines: \textbf{DAgger}, which continues supervised training on aggregated intervention rollouts~\citep{ross2011reduction}; \textbf{AWR}, which reweights chunks using action-level advantages following ALOE~\citep{peng2019advantage,yang2026aloe}; and \textbf{RA-BC}, which reweights data using a reward model following SARM~\citep{chen2025sarm}. Additional details of the policy improvement evaluation are provided in Appendix~\ref{app:C}.

\vspace{-6pt}

\paragraph{Policy improvement through iterative offline post-training.} 
As shown in Fig.~\ref{fig:policy_learning_results}, GR-BC achieves the highest final success rate on all four tasks and improves more consistently across offline post-training rounds. DAgger benefits from additional intervention data but still clones autonomous errors and recovery actions uniformly. AWR is sensitive to inaccurate action-value estimates, which can overweight harmful chunks, while RA-BC is limited by the granularity of its learned progress model when local contact errors or misalignments occur. In contrast, GR-BC weights each executed chunk by its graph-retrieved progress change, suppressing regressive behaviors while prioritizing clean execution and recovery segments. (see Fig.~\ref{fig:progress_model_comparison} for signal-level evidence of this weighting capability).

\vspace{-1.2em}
\subsection{Ablation Study}
\vspace{-6pt}
We ablate graph construction on the Assembly Model dataset to evaluate its effect on graph-retrieved progress. 
\begin{wraptable}{r}{0.54\linewidth}
\vspace{-8pt}
\centering
\caption{Graph ablation on the Assembly Model task.}
\label{tab:graph_ablation}
\scriptsize
\setlength{\tabcolsep}{5pt}
\renewcommand{\arraystretch}{0.9}
\begin{tabular}{@{}lcccc@{}}
\toprule
Variant
& \shortstack{Failure\\reach. $\uparrow$}
& \shortstack{Zero-wt.\\ratio $\downarrow$}
& \shortstack{Pre/\\normal $\downarrow$}
& \shortstack{Mismatch\\edges $\downarrow$} \\
\midrule
Temporal only
& 0.0\%
& 62.3\%
& 1.02
& -- \\
w/o gripper
& 97.8\%
& 26.3\%
& 0.46
& 15.8\% \\
\rowcolor[gray]{0.82}
Ours
& 98.3\%
& 25.0\%
& 0.42
& 0.0\% \\
\bottomrule
\end{tabular}
\vspace{-4pt}
\end{wraptable}
Table~\ref{tab:graph_ablation} reports four diagnostics: failure reachability measures whether failed-rollout nodes can reach the success sink, zero-weight ratio measures how many chunks are suppressed, pre/normal compares weights on pre-intervention autonomous segments against normal segments, and mismatch edges count gripper-inconsistent similarity links. Temporal edges alone cannot connect failed rollouts to the success sink, while cross-trajectory similarity edges greatly improve failure reachability but may introduce incompatible shortcuts. Our gripper-consistent graph removes these mismatches while preserving similar reachability, indicating better graph precision without sacrificing useful recovery connections. We further provide a hyperparameter analysis in Appendix~\ref{app:D}.
\vspace{-8pt}
\section{Limitations}
\vspace{-6pt}
GR-BC is limited by the quality of its dataset-supported reachability graph.
Incorrect visual neighbors can create shortcut paths and distort remaining-step estimates, while overly conservative edge filters may miss valid recovery bridges.
The current gripper-consistency and multi-view checks reduce common mismatch cases, but they do not fully certify physical reachability.
Moreover, shorter graph paths do not necessarily correspond to safer or more robust behaviors, as the current edge costs do not explicitly account for force or collision risk.
In addition, GR-BC has limited generalization to states outside the support of the collected data.
Future work should study stronger reachability tests, risk-aware edge costs, adaptive graph updates, and broader deployment across more embodiments and task families.

\vspace{-8pt}
\section{Conclusion}
\vspace{-6pt}

We introduced GR-BC, an offline iterative framework for VLA post-training. By constructing a task-conditioned goal-reachability graph, we derive Graph-Retrieved Progress (GRP) to estimate minimal remaining steps, enabling consequence-based weighting for action chunks. This approach effectively prioritizes task-advancing segments while reusing recovery behaviors from failures. Empirical results confirm that GRP provides a robust, critic-free progress signal for VLA refinement in precise robotic manipulation.

\clearpage

\acknowledgments{This work was supported in part by the Hangzhou National AI Application Pilot Base (Embodied Intelligence) and the National Natural Science Foundation of China (Grant Nos. 62676327 and 62306242).}

\bibliography{references}

\clearpage

\appendix

\begingroup
\setlist[itemize]{leftmargin=12pt,itemsep=1pt,topsep=2pt}
\setlist[enumerate]{leftmargin=15pt,itemsep=1pt,topsep=2pt}

\newcolumntype{Y}{>{\raggedright\arraybackslash}X}
\newcommand{\graycell}[1]{\cellcolor[gray]{0.90}#1}
\newcommand{\G}{\mathcal{G}}

\section*{Appendix}

\section{Implementation Details and Notation Clarifications}
\label{app:A}
This appendix provides the implementation and evaluation details that are omitted from the main paper due to space. The goal is to make the real-robot experiments reproducible, clarify the data used by each component, and make explicit which information is and is not used when computing graph-retrieved progress (GRP). Unless otherwise stated, all policy-learning experiments use the cumulative offline dataset
\[
  \D^r = \Ddemo \cup \Drollout^{1:r},
  \qquad
  \Drollout^{1:r} = \D_{\mathrm{success\text{-}rollout}}^{1:r}
  \cup \D_{\mathrm{fail\text{-}rollout}}^{1:r},
\]
where rollouts may include human interventions. The graph is built over observed states, while the VLA objective remains a supervised action-chunk prediction loss.

\subsection{Real-World Setup and Task Details}

\begin{wrapfigure}{r}{0.6\linewidth}
  \vspace{-8pt}
  \centering
  \includegraphics[width=\linewidth]{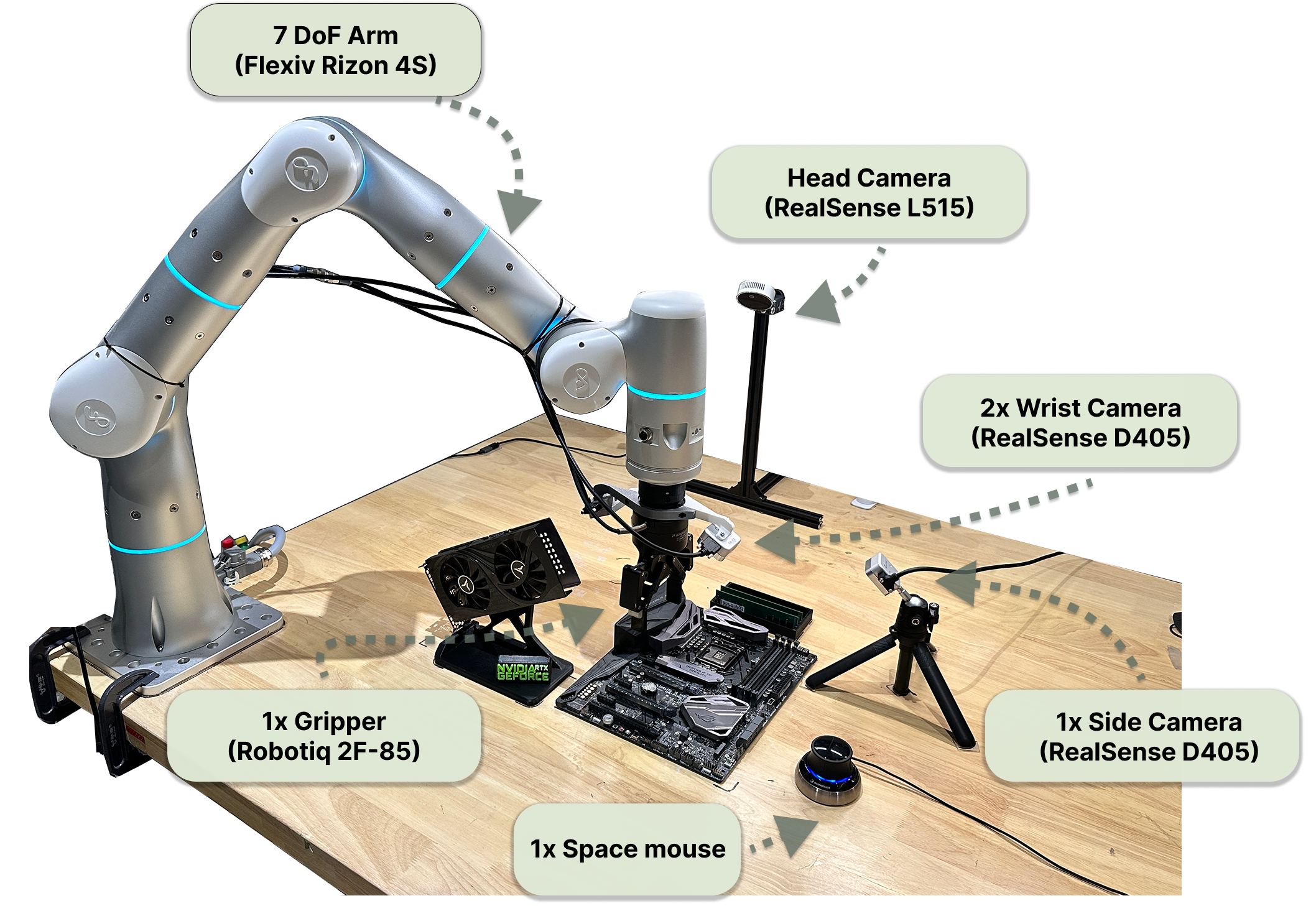}
  \caption{Real-world robot workspace used for GR-BC data collection and evaluation.}
  \label{fig:workspace}
  \vspace{-3pt}
\end{wrapfigure}
Our real-world manipulation platform consists of a 7-DoF Flexiv Rizon 4s arm equipped with a Robotiq 2F-85 parallel gripper, a head-mounted RealSense L515 camera, two wrist-mounted RealSense D405 cameras, one side RealSense D405 camera, and a SpaceMouse used for shared-control interventions. The workspace and camera arrangement are shown in Figure~\ref{fig:workspace}. Although the hardware rig contains four RGB-D cameras, each task uses a fixed subset of three RGB views during data collection and training. This design adapts the observation geometry to the visual requirements of each precise manipulation task without changing the policy interface. Across all tasks, the action space is 8-dimensional, and the robot control loop runs at 30 Hz during both data collection and inference. All real-robot policy deployment and online inference runs are executed on a single NVIDIA GeForce RTX 5090 GPU. We do not apply focal-length scaling in the reported task configurations.

We evaluate GR-BC on four long-horizon manipulation tasks. Each task is judged by terminal physical success rather than dense progress labels. A rollout may therefore be labeled as failed even when it contains recoverable local segments, and a successful demonstration may still contain hesitations, small contact corrections, or local regressions.

\textbf{Assembly Model.}
The robot grasps a black four-hole assembly part and inserts it into a white four-corner socket model. 
The task has broad generalization because the part can start from a large tabletop region. 
It admits two valid success modes, where the robot grasps the part from either the left or right side. 
These modes form multiple successful terminal clusters in the reachability graph; Appendix~\ref{sec:multi_success_sinks} provides the corresponding analysis. 
Success requires full insertion followed by gripper opening.

\textbf{Graphics Card Assembly.}
The robot inserts a graphics card into the target PCIe slot. This task combines large-amplitude arm motion with precise connector alignment~\citep{jing2026assemlm}: the graphics card is relatively heavy, while the bottom latch and gold-finger connector must be inserted accurately. We use an RTX 4060 graphics-card mockup for this task. Success requires the card to be fully inserted and the gripper to open.

\vspace{-2pt}
\begin{table}[H]
  \centering
  \begin{minipage}[t]{0.48\linewidth}
    \vspace{0pt}
    \centering
    \caption{Configuration for Assembly Model.}
    \label{tab:assembly_config}
    \vspace{-4pt}
    \scriptsize
    \setlength{\tabcolsep}{5pt}
    \renewcommand{\arraystretch}{0.9}
    \begin{tabularx}{\linewidth}{@{}p{0.42\linewidth}Y@{}}
      \toprule
      Field & Setting \\
      \midrule
      Task instruction & Pick up the black connector and insert it into the white socket. \\
      Camera configuration & Head, left wrist, right wrist; no focal-length scaling \\
      Initial demonstrations & 215 \\
      Mean episode length & 376 \\
      Action chunk horizon \(N\) & 30 frames \\
      Graph stride & \(k=15\) frames \\
      \bottomrule
    \end{tabularx}
  \end{minipage}
  \hfill
  \begin{minipage}[t]{0.48\linewidth}
    \vspace{0pt}
    \centering
    \caption{Configuration for Graphics Card.}
    \label{tab:graphics_config}
    \vspace{-4pt}
    \scriptsize
    \setlength{\tabcolsep}{5pt}
    \renewcommand{\arraystretch}{0.9}
    \begin{tabularx}{\linewidth}{@{}p{0.42\linewidth}Y@{}}
      \toprule
      Field & Setting \\
      \midrule
      Task instruction & Pick up the graphics card and insert it into the motherboard PCIe slot. \\
      Camera configuration & Head, left wrist, side; no focal-length scaling \\
      Initial demonstrations & 125 \\
      Mean episode length & 911 \\
      Action chunk horizon \(N\) & 30 frames \\
      Graph stride & \(k=15\) frames \\
      \bottomrule
    \end{tabularx}
  \end{minipage}
\end{table}
\vspace{-5pt}

\textbf{Screw Tightening.}
The robot uses a screwdriver to tighten two screws sequentially. This is a long-horizon and contact-rich task: before rotation, the policy must localize the small screw head and align the tool tip accurately. The task uses M2 screw heads. Success requires each of the two screws to be tightened by 90 degrees.

\textbf{Memory Stick Insertion.}
The robot picks up a thin memory stick from a box and inserts it into the target socket. The task requires generalization over randomized object poses and demands finer connector alignment than the graphics-card assembly task. We use a DDR4 memory stick, whose socket opening is approximately 1 mm wide. Success requires the memory stick to be fully inserted and the slot latch to lock.

\vspace{0pt}
\begin{table}[H]
  \centering
  \begin{minipage}[t]{0.48\linewidth}
    \vspace{0pt}
    \centering
    \caption{Configuration for Screw Tightening.}
    \label{tab:screw_config}
    \vspace{-3pt}
    \scriptsize
    \setlength{\tabcolsep}{5pt}
    \renewcommand{\arraystretch}{0.9}
    \begin{tabularx}{\linewidth}{@{}p{0.42\linewidth}Y@{}}
      \toprule
      Field & Setting \\
      \midrule
      Task instruction & Pick up the screwdriver, tighten the two screws one after the other. \\
      Camera configuration & Head, right wrist, side; 3x focal-length scaling \\
      Initial demonstrations & 100 \\
      Mean episode length & 1,517 \\
      Action chunk horizon \(N\) & 30 frames \\
      Graph stride & \(k=15\) frames \\
      \bottomrule
    \end{tabularx}
  \end{minipage}
  \hfill
  \begin{minipage}[t]{0.48\linewidth}
    \vspace{0pt}
    \centering
    \caption{Configuration for Memory Stick.}
    \label{tab:memory_config}
    \vspace{-3pt}
    \scriptsize
    \setlength{\tabcolsep}{5pt}
    \renewcommand{\arraystretch}{0.9}
    \begin{tabularx}{\linewidth}{@{}p{0.42\linewidth}Y@{}}
      \toprule
      Field & Setting \\
      \midrule
      Task instruction & Pick up the RAM and insert it into the motherboard RAM slot. \\
      Camera configuration & Left wrist, right wrist, side; no focal-length scaling \\
      Initial demonstrations & 150 \\
      Mean episode length & 1,008 \\
      Action chunk horizon \(N\) & 30 frames \\
      Graph stride & \(k=15\) frames \\
      \bottomrule
    \end{tabularx}
  \end{minipage}
\end{table}
\vspace{-5pt}

Rollouts are collected with SpaceMouse-based shared control. During policy execution, any nonzero SpaceMouse input triggers human takeover. Control is returned to the policy after a short fixed timeout with no SpaceMouse input. The resulting trajectories interleave autonomous execution, failure-prone behavior, and corrective segments, making them useful for offline data analysis, policy post-training, and reward-model evaluation pipelines with different input requirements.
\vspace{-3pt}
\subsection{Dataset Composition}
\vspace{-3pt}

The initial dataset contains expert demonstrations~\citep{jing2026humanoidgen}. In each offline iteration, the current policy is deployed to collect additional real-world rollouts, which are retained regardless of final outcome. Table~\ref{tab:dataset_stats} summarizes the number of demonstration trajectories, successful and failed rollouts, and the mean number of rollouts collected per offline round for each task.

\begin{table}[t]
  \centering
  \caption{Aggregate dataset statistics.}
  \label{tab:dataset_stats}
  \small
  \setlength{\tabcolsep}{4pt}
  \begin{tabularx}{\linewidth}{@{}lYYYY@{}}
    \toprule
    Statistic & Assembly Model & Graphics Card & Screw Tightening & Memory Stick \\
    \midrule
    Demo trajectories & 215 & 125 & 100 & 150 \\
    Successful rollouts & 85 & 270 & 160 & 240 \\
    Failed rollouts & 100 & 120 & 80 & 80 \\
    Mean rollouts per round & 90 & 120 & 80 & 100 \\
    \bottomrule
  \end{tabularx}
\end{table}
\vspace{-5pt}

\subsection{Graph Construction Details}

For each task, we construct a task-conditioned reachability graph \(\G=(\mathcal{V},\mathcal{E})\) over the cumulative offline dataset. Nodes are subsampled trajectory states, with all episode terminal states and all successful terminal states retained. Each node stores synchronized multi-view visual embeddings, proprioception, and the binary gripper state. The visual encoder is the manipulation-centric representation (MCR) model \citep{jiang2025robots}; we compare this choice against R3M \citep{nair2022r3m} and V-JEPA2.1 variants in Appendix~\ref{sec:encoder_comparison}.
For Screw Tightening, we incorporate historical frames into each camera's embedding to provide temporal context for the repetitive motions involved in tightening two screws sequentially.

\begin{wraptable}{r}{0.50\linewidth}
  \vspace{-10pt}
  \centering
  \caption{Hyperparameters mentioned in the paper.}
  \label{tab:graph_hparams}
  \vspace{2pt}
  \scriptsize
  \setlength{\tabcolsep}{2pt}
  \renewcommand{\arraystretch}{0.88}
  \begin{tabularx}{\linewidth}{@{}p{0.15\linewidth}p{0.53\linewidth}Y@{}}
    \toprule
    Parameter & Meaning & Default \\
    \midrule
    \(k\) & temporal subsampling interval & 15 frames \\
    \(K\) & mutual KNN size per camera view & 20 \\
    \(\alpha_{\mathrm{edge}}\) & adaptive similarity-edge threshold & 1.0 \\
    \(\epsilon_c\) & stability constant for adaptive length & \(10^{-4}\) \\
    \(N\) & action chunk horizon & 30 frames \\
    \(\delta_{\min}\) & minimum net progress threshold & 0.002 \\
    \(\eta\) & maximum allowed progress regression & 0.03 \\
    \(\beta_w\) & progress bonus amplitude & 0.5 \\
    \(\gamma\) & progress sensitivity & 20 \\
    \(w_{\min}\) & lower accepted-branch weight & 0.5 \\
    \(w_{\max}\) & upper accepted-branch weight & 1.5 \\
    \bottomrule
  \end{tabularx}
  \vspace{-2pt}
\end{wraptable}

Temporal edges encode physically executed transitions within a trajectory. Similarity edges bridge states from different trajectories when they pass multi-view mutual-nearest-neighbor retrieval, gripper-state consistency, and adaptive edge-length thresholding. Proprioception is stored for diagnostics and policy input but is not used in similarity retrieval because the multi-view MCR features plus gripper-state filtering provided the most stable retrieval interface in our scans. Accepted similarity pairs are implemented as two directed edges with identical cost so that reverse shortest-path search can be run on a directed graph. Similarity edges are not interpreted as executable transitions; they only identify visually/proprioceptively compatible states for potential estimation. We connect all successful terminal nodes to the virtual sink \(g_\star\), then run reverse Dijkstra search \citep{dijkstra2022note} to compute \(\Rrec(v)\), the minimum graph cost from \(v\) to any successful terminal state. Unless otherwise specified, we use the 95th percentile of finite \(\Rrec\) as \(R_{\max}\). When converting sparse node progress to dense frame-level progress, we interpolate only finite node pairs; otherwise progress is set to 0.

Figure~\ref{fig:assembly_two_traj_graph} illustrates the graph construction process on the Assembly Model task using two representative trajectories. The left panel shows temporally connected trajectory nodes, cross-trajectory similarity links, the virtual sink \(g_\star\), and the retrieved remaining-step values \(\Rrec(v)\). Here \(R\) denotes the graph-retrieved remaining-step estimate from the current node to a successful terminal node, so smaller values indicate states closer to success and \(R=0\) corresponds to a successful terminal node or the virtual success sink. The right panel visualizes matched image states used to verify the local visual consistency of the accepted similarity edges; \(d\) denotes the MCR feature-space distance between the two matched camera-image states, where smaller values indicate stronger visual or representational similarity.

Figure~\ref{fig:graph_overview} visualizes the resulting graphs built from the demonstration data and one rollout round across all four tasks.

\begin{figure}[t]
  \centering
  \includegraphics[width=\linewidth]{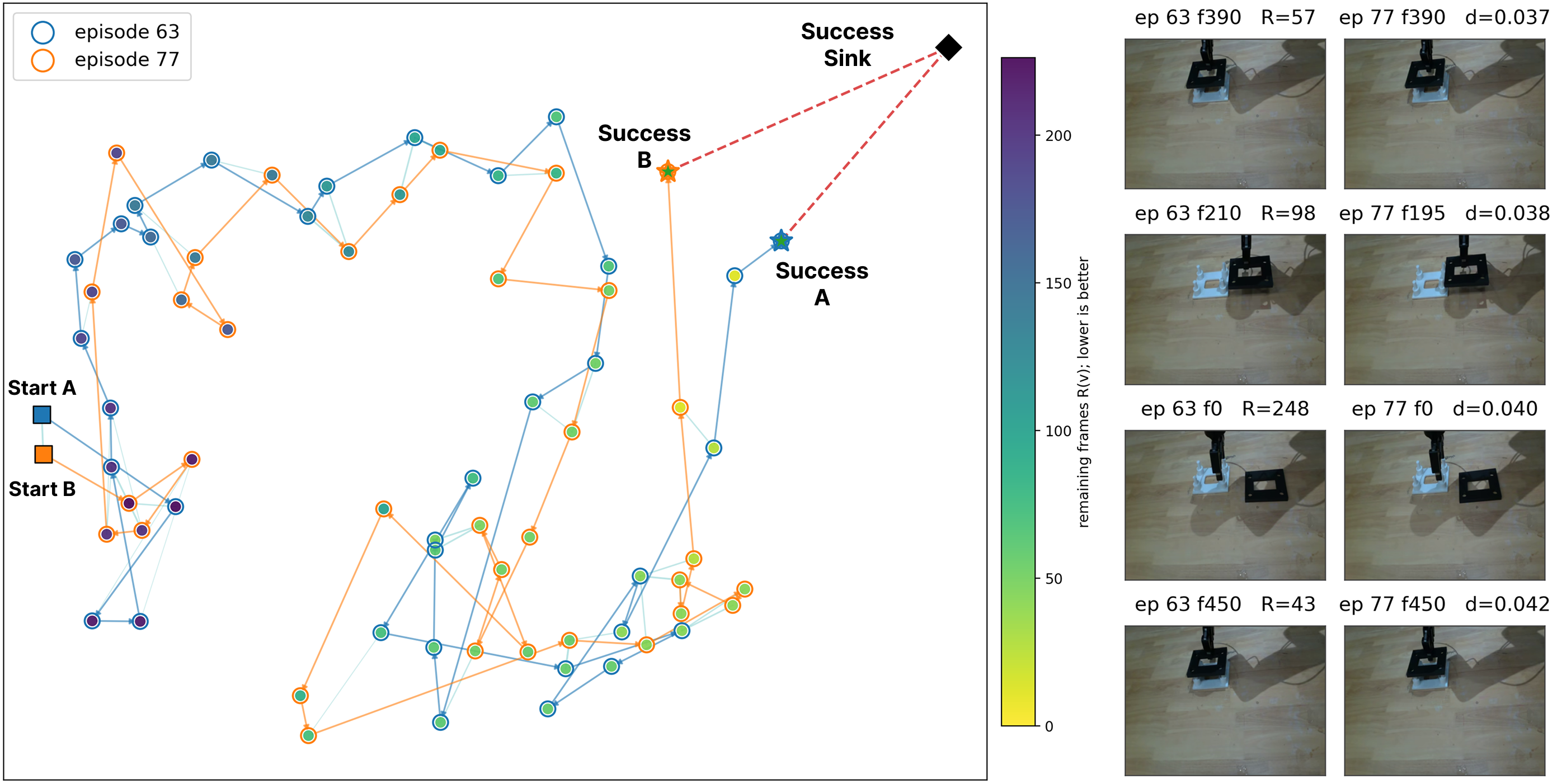}
  \caption{Two-trajectory graph construction example on the Assembly Model task. Blue and orange nodes denote two different trajectories, solid edges denote temporal transitions, and dashed red edges denote zero-cost edges from successful terminal nodes to \(g_\star\). Node colors indicate retrieved remaining steps \(\Rrec(v)\), where lower values correspond to states closer to task completion. The right image pairs show representative matched states; \(R\) is the graph-retrieved remaining-step estimate and \(d\) is the MCR feature-space distance between the matched states.}
  \label{fig:assembly_two_traj_graph}
\end{figure}

\begin{figure}[H]
  \centering
  \includegraphics[width=\linewidth]{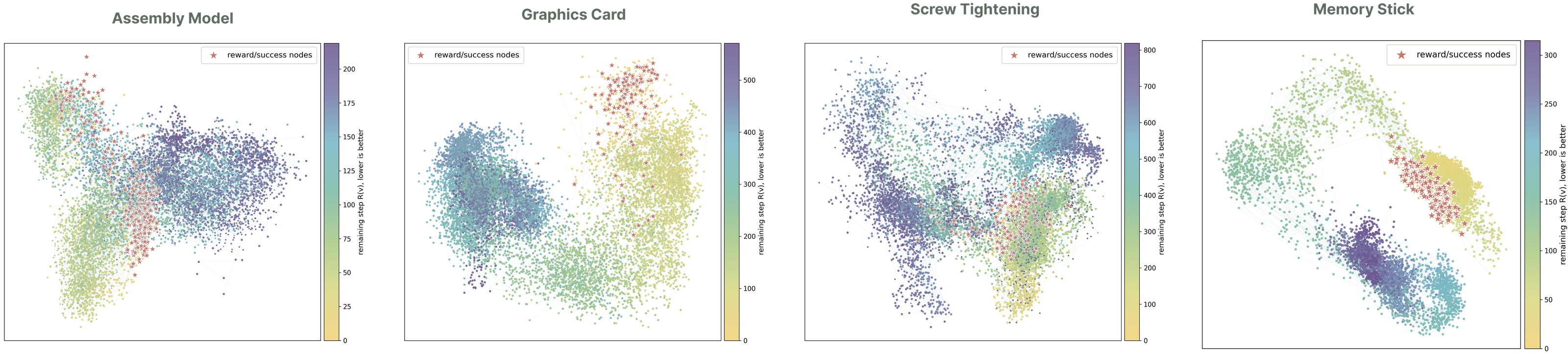}
  \caption{Graph overview constructed from the demonstration data and one rollout round for all four tasks. Darker node colors indicate larger remaining-step estimates.}
  \label{fig:graph_overview}
\end{figure}

\section{Progress Evaluation Protocol and Additional Progress Results}
\label{app:B}

This section provides the progress-evaluation protocol used to compare GRP with learned or foundation-model-based progress and reward baselines, including ReWiND \citep{zhang2025rewind}, SARM \citep{chen2025sarm}, Robo-Dopamine \citep{tan2025robo}, and RoboMeter \citep{liang2026robometer}. These methods differ in their intended supervision and data interfaces; therefore, we report both interface assumptions and fixed evaluation metrics instead of forcing all methods into an artificial identical training protocol.

\subsection{Baseline Implementation and Supervision}

The progress baselines are evaluated under their intended supervision settings. ReWiND uses language-guided reward estimation \citep{zhang2025rewind}; SARM learns stage-aware rewards from stage or progress supervision \citep{chen2025sarm}; Robo-Dopamine and RoboMeter provide general-purpose process or trajectory-comparison reward signals \citep{tan2025robo,liang2026robometer}. These learned reward baselines are therefore not supervision-matched to GRP: they may use progress labels, segment comparisons, or failure-derived examples when required by their own training interfaces. GRP instead uses terminal success information and graph connectivity, without dense progress labels, subtask labels, or trajectory comparisons.

Across the four real-robot tasks, we use the same evaluation protocol and keep each learned progress baseline as close as possible to its intended training recipe. All progress pipelines in the baseline comparison are instantiated separately for each task, not only for Screw Tightening. For each task, progress metrics are computed on a fixed evaluation set with 30 successful rollout episodes and 30 failed rollout episodes from that task, and the reported baseline comparisons summarize the four-task evaluation rather than a single Screw Tightening run. In the main-paper progress table, \(\mathrm{Class.\ SE}\) and \(\mathrm{Class.\ FE}\) are averaged over the four tasks and reported out of 30 episodes per task. For GRP, this primary progress table is a \emph{transductive graph diagnostic}: the evaluation trajectories are scored within the task graph to test whether graph-retrieved progress separates known successful and failed trajectory structure. It should not be interpreted as held-out progress generalization. The held-out query setting is evaluated separately in Appendix~\ref{app:demo_only_zero_shot}, where held-out rollouts are inserted only as query trajectories and their terminal outcomes are not used to define success sinks. When a baseline requires task-specific training or adaptation, we train it for approximately 20 epochs whenever the method implementation and compute budget permit, and otherwise use the closest method-prescribed schedule. We avoid introducing a separate hand-tuned training protocol for each task; method-specific supervision, adaptation, and input-interface details are described below for each baseline.

\paragraph{ReWiND.}
ReWiND trains a language-guided Transformer reward model to predict task progress from trajectory prefixes. It uses DINOv2 visual embeddings and MiniLM language embeddings, with forward progress supervision, rewound segments, and mismatched-language negatives. For each task, we run the ReWiND pipeline with that task's language instruction and trajectories, using the official training code while preserving its language-conditioned reward-estimation interface.

\paragraph{SARM.} 
SARM learns a language-conditioned progress reward from robot trajectory videos, avoiding hand-designed dense rewards. For each task, we run SARM with that task's language instruction and trajectories. We use its \textbf{single-stage mode}, where each successful trajectory is treated as one continuous task stage. We intentionally use the strongest setting that does not require extra stage annotation, rather than providing manually segmented stage labels for these long-horizon precise manipulation tasks. Frozen CLIP encoders extract visual and language embeddings, and a Transformer reward head predicts normalized task progress over temporal windows. For efficiency, we precompute CLIP embeddings and evaluate the model using dense per-frame progress curves.

\paragraph{Robo-Dopamine.}
Robo-Dopamine learns a general process reward model from language instructions
and multi-view visual comparisons. Given reference start/end observations and a
pair of BEFORE/AFTER observations, the model predicts relative task progress or
regression. For each task, we adapt the released Robo-Dopamine GRM
checkpoint with LoRA on converted LeRobot trajectories. Each training example
contains the task instruction, reference start/end images, and sampled
BEFORE/AFTER multi-view images. Successful trajectories provide temporal
progress-change supervision. We do not use failed trajectories in the Robo-Dopamine
adaptation; its \emph{Failure data} entry is therefore marked as ``--'' in the
baseline-interface summary. These progress-change labels are part of the Robo-Dopamine
baseline interface and are counted as additional progress supervision,
not as information used by GRP. We freeze the multimodal backbone and train only
task-specific LoRA adapters, and evaluate the adapted GRM by converting its
predicted progress changes into dense per-frame progress curves.

\paragraph{RoboMeter.}
RoboMeter is a general-purpose vision-language reward model that combines
progress, success, and trajectory-preference prediction. It builds on a
Qwen3-VL backbone with separate reward heads for progress, preference, and
success. For each task, we fine-tune the RoboMeter-4B checkpoint on
the corresponding processed task trajectories. Terminal rewards provide trajectory-level
success labels, while sampled trajectory segments are used to construct
preference examples that also supervise the progress and success heads. These
preference and terminal-success signals are part of the RoboMeter baseline
interface and are counted as additional reward-model supervision, not as
information used by GRP. In our implementation, this baseline uses the wrist-view camera as its visual input. We
evaluate the resulting reward model using its predicted progress and success
scores as dense per-frame reward signals.

\subsection{Metric Definitions}

For an episode \(\tau\), let \(p(s_t)\in[0,1]\) denote the normalized progress estimate at time \(t\). We report four metrics.

\paragraph{\(\boldsymbol{\Delta}_{\mathrm{SF}}\).}
The success--failure progress gap \(\Delta_{\mathrm{SF}}\) measures separation between successful and failed episodes:
\[
  \Delta_{\mathrm{SF}}
  =
  \mathbb{E}_{\tau\in \mathrm{SE}}\left[p(s_T^\tau)\right]
  -
  \mathbb{E}_{\tau\in \mathrm{FE}}\left[p(s_T^\tau)\right],
\]
where \(s_T^\tau\) is the terminal state of trajectory \(\tau\). Larger values indicate better final-outcome separation.

\paragraph{\(\boldsymbol{\rho}_{\mathrm{dev}}\).}
We manually localize a fixed set \(W_{\mathrm{dev}}\) of deviation windows that contain visible local regressions, such as teleoperation mistakes, contact misalignment, or pre-intervention deviations. The deviation-detection rate is
\[
  \rho_{\mathrm{dev}}
  =
  \frac{1}{|W_{\mathrm{dev}}|}
  \sum_{(t_s,t_e)\in W_{\mathrm{dev}}}
  \ind\left[p(s_{t_s})-p(s_{t_e})>\delta_{\mathrm{dev}}\right].
\]
Higher \(\rho_{\mathrm{dev}}\) indicates greater sensitivity to fine-grained local regressions.

\paragraph{Deviation-window annotation protocol.}
The deviation windows are annotated before computing any method's progress scores. For each of the four tasks, an annotator inspects that task's fixed evaluation rollout videos and marks short intervals where the robot visibly moves away from task completion, loses contact alignment, executes an incorrect teleoperation correction, or enters the frames immediately preceding an intervention. The annotator uses only the raw rollout video and task context, not GRP scores, learned-reward scores, model predictions, or any progress curve. Each annotation stores the task name, episode index, start frame, end frame, and the label \texttt{bad\_interval}; the same frozen per-task annotation files are then used for all methods. We set \(\delta_{\mathrm{dev}}\) once on the normalized progress scale and do not tune it per method or task. This makes \(\rho_{\mathrm{dev}}\) a blind, window-level test of whether a progress estimator decreases over human-identified local regressions, rather than a window set selected from a model's output.

\paragraph{\(\mathbf{Class.\ SE}\) and \(\mathbf{Class.\ FE}\).}
Following the automated judgment logic of SARM, we classify each evaluation episode \(\tau\) using its predicted progress sequence \(p(s_t^\tau)\). The predicted episode label is defined as
\[
\mathrm{Label}=
\begin{cases}
\mathrm{SE}, &
\text{if } p(s_T^\tau) > 0.8
\ \wedge\
\frac{3}{T}\sum_{t=\lceil 2T/3\rceil}^{T} p(s_t^\tau) > 0.6, \\[3pt]
\mathrm{FE}, &
\text{otherwise},
\end{cases}
\]

\(\mathrm{Class.\ SE}\) measures how many successful episodes are correctly recognized as successful, while \(\mathrm{Class.\ FE}\) measures how many failed episodes are correctly rejected as failures.

\subsection{Demo-Only Zero-Shot Progress Evaluation}
\label{app:demo_only_zero_shot}

\begin{wrapfigure}{r}{0.54\linewidth}
  \vspace{-8pt}
  \centering
  \includegraphics[width=\linewidth]{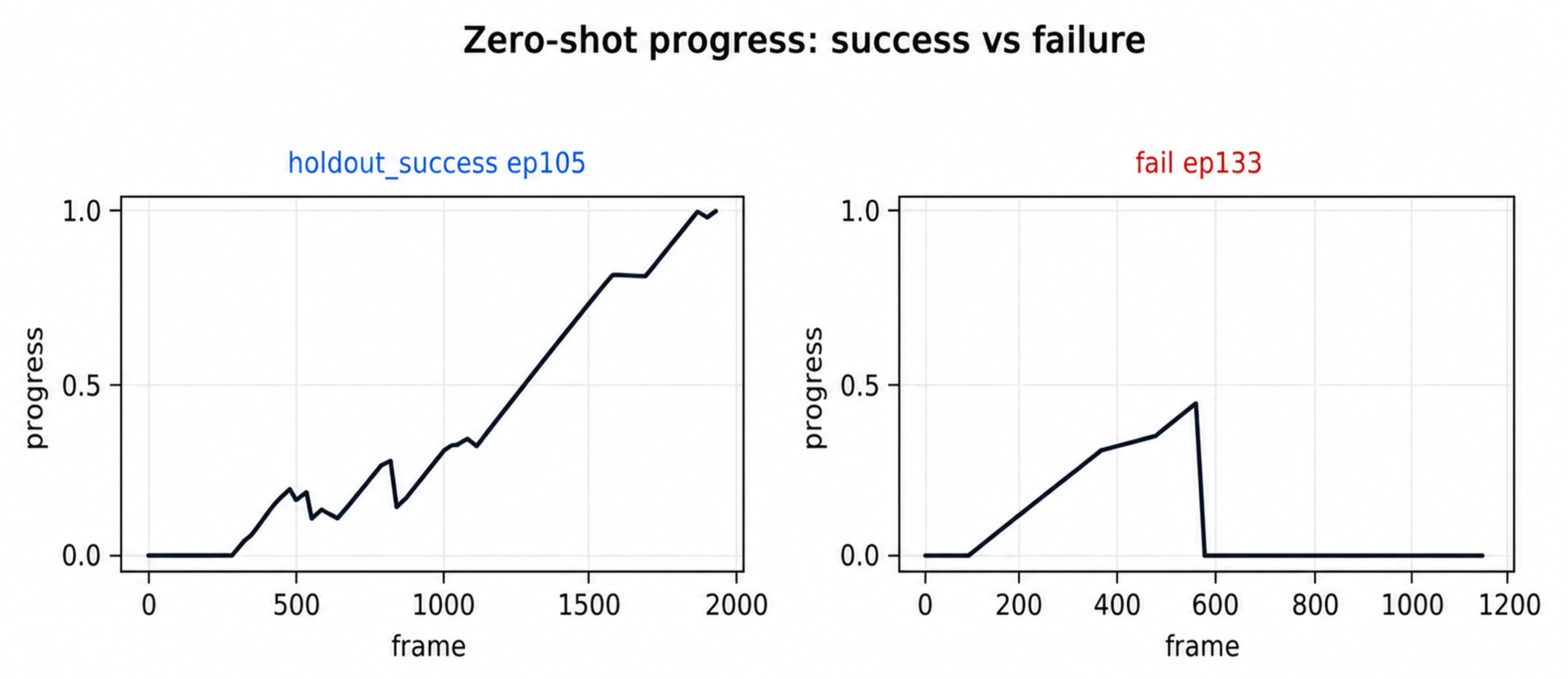}
  \caption{Demo-only zero-shot progress evaluation on held-out Screw Tightening rollouts.}
  \label{fig:zeroshot_progress}
  \vspace{-2pt}
\end{wrapfigure}

As an additional single-task diagnostic, separate from the four-task baseline comparison above, we evaluate a stricter zero-shot progress setting that approximates online rollout scoring before rollout aggregation by using a Screw Tightening dataset.
The base graph is built only from 100 successful expert demonstrations.
We then reserve 18 additional successful trajectories and 17 failed trajectories as held-out rollouts.
Each held-out rollout is temporarily inserted into the demonstration-only graph, scored by graph-retrieved progress (GRP), and removed before the next rollout is evaluated.
Thus, held-out rollouts cannot exchange information with each other, and their outcomes are never used to define success states.

This protocol tests whether GRP can provide a query-time progress signal for unseen online rollouts before they are added to the training pool.
Table~\ref{tab:demo_only_zero_shot} reports trajectory-level progress diagnostics.
\textbf{Terminal Progress} is the progress value at the final sampled state of a rollout.
\textbf{Trajectory Avg. Progress} averages progress over all sampled states in the rollout.
\textbf{Low-Progress State Ratio} measures the fraction of sampled states whose progress is below the low-progress threshold, indicating how often the rollout stays far from the demonstrated success manifold.
\textbf{Reachable State Ratio} measures the fraction of sampled states that can reach a demonstration success node through the graph.
\textbf{False Classification} reports false negatives (FN) for held-out successful rollouts and false positives (FP) for failed rollouts.

In this diagnostic split, GRP yields a large terminal-progress gap between held-out successful and failed rollouts.
All failed trajectories receive zero terminal progress, while most held-out successful trajectories reach high terminal progress.
This suggests that graph-retrieved progress can provide a useful query-time rollout score in this setting.
As verified online rollouts are appended to the graph, the reachable state support expands, so the progress signal can become increasingly accurate without retraining a dense neural reward model.

\begin{table}[ht]
\centering
\caption{\textbf{Demo-only zero-shot progress evaluation.}
The graph is built from successful demonstrations only.
Held-out rollouts are inserted one at a time, scored by graph-retrieved progress, and removed before the next rollout is evaluated.
The diagnostic split yields a large terminal-progress gap between held-out successful and failed rollouts.}
\label{tab:demo_only_zero_shot}
\scriptsize
\setlength{\tabcolsep}{4pt}
\renewcommand{\arraystretch}{0.95}
\begin{tabular}{lcccccc}
\toprule
Split
& \# Traj.
& \shortstack{Terminal\\Progress}
& \shortstack{Trajectory\\Avg. Progress}
& \shortstack{Low-Progress\\State Ratio}
& \shortstack{Reachable\\State Ratio}
& \shortstack{False\\Classification} \\
\midrule
Held-out success
& 18
& $0.879$
& $0.440$
& $0.328$
& $0.998$
& $2/18$ FN \\
Held-out failure
& 17
& $\mathbf{0.000}$
& $\mathbf{0.019}$
& $\mathbf{0.952}$
& $0.127$
& $\mathbf{0/17}$ FP \\
\rowcolor[gray]{0.90}
Success--failure gap
& --
& $\mathbf{0.879}$
& $\mathbf{0.421}$
& $\mathbf{-0.624}$
& $\mathbf{0.871}$
& -- \\
\bottomrule
\end{tabular}
\vspace{2pt}
\begin{flushleft}
\scriptsize
\emph{Note:} The demo-only base graph contains 8756 nodes, 8656 temporal edges, 9700 similarity edges, and 100 successful terminal nodes.
\end{flushleft}
\end{table}

\section{Policy Improvement Evaluation Details and Single-task Backbone Compatibility Study}
\label{app:C}

This section provides the raw policy-evaluation protocol, baseline training details, and an additional controlled single-task backbone compatibility study for using GR-BC as a post-training interface beyond the X-VLA setting evaluated in the main four-task study.

\subsection{VLA Training Details}

The main iterative policy-learning study uses X-VLA \citep{zheng2025x} because it enables controlled multi-round real-robot evaluation under the same data and compute budget. The single-task backbone compatibility study in Appendix~\ref{sec:backbone_generality} additionally evaluates SmolVLA \citep{shukor2025smolvla} and \(\pi_{0.5}\) \citep{black2025pi_}. Across backbones, GR-BC preserves the original supervised VLA loss and only changes the sample-level multiplier \(w_t\).

\begin{table}[t]
  \centering
  \caption{VLA post-training hyperparameters.}
  \label{tab:vla_hparams}
  \small
  \resizebox{\linewidth}{!}{
  \begin{tabular}{llll}
    \toprule
    Parameter & X-VLA & SmolVLA & \(\pi_{0.5}\) \\
    \midrule
    Input image resolution
      & \makecell[l]{Raw videos: \(480\times640\);\\ model inputs: \(256\times256\), \\ \(256\times256\), \(224\times224\)}
      & \makecell[l]{\(480\times640\) raw RGB videos;\\ padded to \(512\times512\)}
      & \makecell[l]{\(480\times640\) raw RGB videos;\\ resized to \(224\times224\)} \\
    Number of camera views
      & 3 & 3 & 3 \\
    Action representation
      & \makecell[l]{8-D joint/gripper\\ command}
      & \makecell[l]{8-D joint/gripper\\ command}
      & \makecell[l]{8-D joint/gripper\\ command} \\
    Action chunk horizon \(N\)
      & 30 frames
      & 50 frames
      & 30 frames \\
    Hardware
      & \makecell[l]{4 NVIDIA H100 GPUs}
      & \makecell[l]{4 NVIDIA H100 GPUs}
      & \makecell[l]{4 NVIDIA H100 GPUs} \\
    Batch size
      & \makecell[l]{32 per GPU;\\ 128 effective}
      & \makecell[l]{32 per GPU;\\ 128 effective}
      & \makecell[l]{16 per GPU;\\ 64 effective} \\
    Learning rate
      & \(1.0\times10^{-4}\)
      & \(1.0\times10^{-4}\)
      & \(2.5\times10^{-5}\) \\
    Optimizer
      & \makecell[l]{\(\beta=(0.9,0.95)\);\\ wd \(=1.0\times10^{-4}\)}
      & \makecell[l]{\(\beta=(0.9,0.95)\);\\ wd \(=1.0\times10^{-10}\)}
      & \makecell[l]{\(\beta=(0.9,0.95)\);\\ wd \(=1.0\times10^{-2}\)} \\
    Training steps
      & 20k
      & 20k
      & 40k \\
    Training time
      & 8h
      & 6h
      & 20h \\
    \bottomrule
  \end{tabular}}
\end{table}

\subsection{Single-task Backbone Compatibility Study on Graphics Card}
\label{sec:backbone_generality}

The full iterative policy-learning comparison in Sec.~5.2 of the main paper uses X-VLA~\citep{zheng2025x} for all baselines, since X-VLA allows controlled multi-round evaluation across four real-world tasks under the same compute and data budget. To test backbone compatibility in a controlled single-task study, we additionally evaluate the Graphics Card task with three different VLA backbones: SmolVLA~\citep{shukor2025smolvla}, \(\pi_{0.5}\)~\citep{black2025pi_}, and X-VLA~\citep{zheng2025x}.

In this study, we compare GR-BC with the DAgger-style uniform aggregation baseline under each backbone. The graph construction, GRP computation, transition weights, data split, and evaluation poses are kept fixed, while only the VLA backbone and its corresponding supervised training objective are changed. For SmolVLA, which predicts a 50-frame action chunk, we use the same 30-frame GRP consequence horizon for weighting and apply the resulting sample-level weight to the full supervised chunk. This preserves a fixed graph-scoring interface while leaving the backbone-specific action objective unchanged. This setting supports backbone compatibility in a controlled single-task study, while leaving broader architecture and task coverage to future evaluation.

\begin{figure}[t]
\centering
\includegraphics[width=\linewidth]{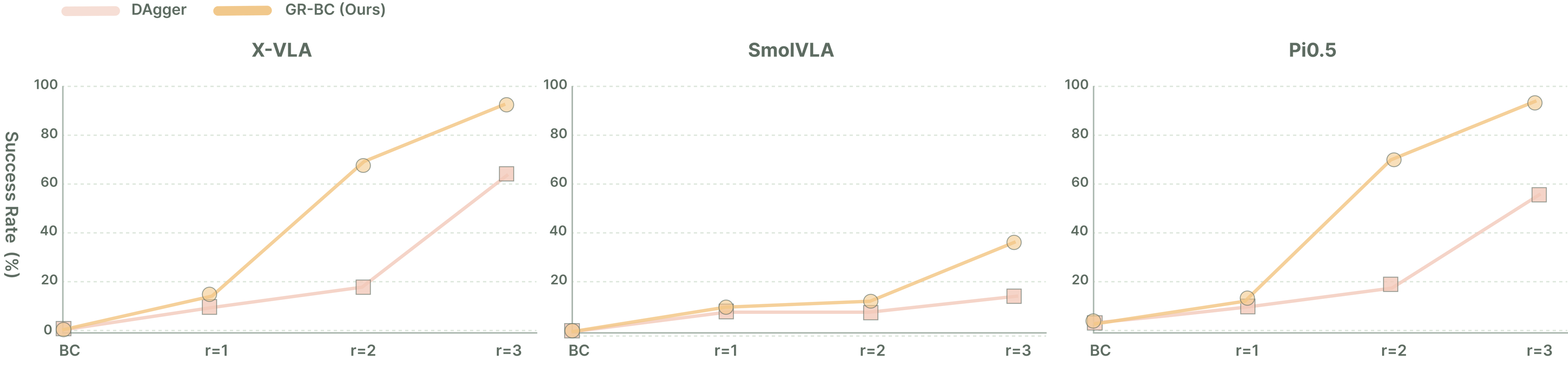}
\caption{Single-task backbone compatibility on Graphics Card. GR-BC improves over the DAgger-style baseline across X-VLA, SmolVLA, and \(\pi_{0.5}\) in this controlled study.}
\label{fig:backbone_generality}
\end{figure}

As shown in Table~\ref{tab:backbone_generality} and Fig.~\ref{fig:backbone_generality}, GR-BC improves over the DAgger-style baseline across all three VLA backbones on the Graphics Card task. The improvement is especially clear after the second and third post-training rounds, where graph-based consequence reweighting better exploits useful recovery and progress-making chunks from the aggregated rollout data. These results support backbone compatibility in a controlled single-task study. Because GR-BC only multiplies the original supervised VLA loss by the precomputed chunk weight \(w_t\), it can be applied to several supervised VLA objectives in this setting, including standard behavior cloning, diffusion-style action prediction~\citep{chi2025diffusion}, and flow-style action prediction.

\subsection{Policy Baseline Training Details}

All policy baselines start from the same BC checkpoint trained on successful demonstrations. The DAgger-style baseline aggregates shared-control rollout data and applies unweighted supervised training \citep{ross2011reduction}; AWR uses action-value-derived weights \citep{peng2019advantage}; RA-BC follows reward-weighted behavior cloning using a learned progress or reward model such as SARM \citep{chen2025sarm}; GR-BC uses graph-retrieved progress and does not train an extra neural reward, value, critic, or progress model. The AWR baseline is conceptually related to advantage-weighted and action-level off-policy VLA post-training methods \citep{peng2019advantage,yang2026aloe}.

\begin{wraptable}{r}{0.50\linewidth}
\centering
\caption{Single-task backbone compatibility study on the Graphics Card task. We report success rate (\%) across behavior cloning initialization and three iterative post-training rounds.}
\label{tab:backbone_generality}
\scriptsize
\setlength{\tabcolsep}{4pt}
\renewcommand{\arraystretch}{1.08}
\begin{tabular}{@{}llcccc@{}}
\toprule
Backbone & Method & BC & \(r=1\) & \(r=2\) & \(r=3\) \\
\midrule
\multirow{2}{*}{X-VLA}
& DAgger-style & 0 & 8 & 17 & 63 \\
& \graycell{GR-BC} & \graycell{0} & \graycell{14} & \graycell{67} & \graycell{92} \\
\midrule
\multirow{2}{*}{SmolVLA}
& DAgger-style & 0 & 8 & 8 & 14 \\
& \graycell{GR-BC} & \graycell{0} & \graycell{10} & \graycell{12} & \graycell{35} \\
\midrule
\multirow{2}{*}{\(\pi_{0.5}\)}
& DAgger-style & 3 & 12 & 18 & 55 \\
& \graycell{GR-BC} & \graycell{3} & \graycell{14} & \graycell{69} & \graycell{93} \\
\bottomrule
\end{tabular}
\end{wraptable}

\paragraph{DAgger-style.}
We implement a DAgger-style uniform aggregation baseline inspired by Dataset Aggregation~\citep{ross2011reduction}, adapted to our shared-control action-chunk setting. This is not a pure oracle-relabeling DAgger setup: we do not retroactively query an expert action for every learner-visited autonomous state. In each offline post-training round, rollout trajectories collected with SpaceMouse shared control are added to the dataset. Autonomous chunks use the action chunk actually executed by the policy, while intervention chunks use the human teleoperation or corrective action chunk. The new data is uniformly aggregated with the historical dataset $\mathcal{D}^r$, and the policy is updated via the standard unweighted maximum-likelihood objective:
\[
\mathcal{L}_{\mathrm{DAgger\text{-}style}} = \mathbb{E}_{(s_t,\, g,\, a_{t:t+N-1}) \sim \mathcal{D}^r} \left[ \mathcal{L}_{\mathrm{VLA}}\!\left(\pi_\phi,\, s_t,\, g,\, a_{t:t+N-1}\right) \right].
\]
Unlike GR-BC, the DAgger-style baseline treats all aggregated transitions as equally positive supervision ($w_t = 1$), meaning it does not evaluate the consequence of individual action chunks or mathematically distinguish between autonomous error segments and expert recovery segments.

\paragraph{RA-BC.}
We implement Reward-Aligned Behavior Cloning (RA-BC) following the pipeline of SARM~\citep{chen2025sarm}, which leverages a learned neural progress model to reweight imitation data. Since none of the compared progress pipelines uses multi-stage task segmentation, we use the same single-stage/single-mode setting for RA-BC: each successful trajectory is treated as one continuous stage, and its progress target is the normalized temporal position within that successful trajectory. We train \(P_\theta(s)\in[0,1]\) on successful demonstrations only, rather than adding rollout, failure, or intervention labels to the reward model. This avoids giving the learned reward baseline extra supervision beyond the single-stage progress interface; rollout and intervention data are still used in the downstream policy dataset, but not as reward-model labels. For each action chunk \(a_{t:t+N-1}\) executed at state \(s_t\), the trajectory impact is evaluated via the endpoint difference
\[
\Delta \hat{p}_t = P_\theta(s_{t+N}) - P_\theta(s_t).
\]
In the LeRobot RA-BC loader, the default endpoint mode uses \(N\) frames ahead within the same episode, falls back to the last frame near episode end, and maps the progress delta \(r_i=\Delta\hat{p}_t\) into a sample weight using \(\kappa=0.01\) and \(\epsilon=10^{-6}\):
\[
\tilde{w}_i=\mathrm{clip}\!\left(\frac{r_i-(\mu-2\sigma)}{4\sigma+\epsilon},0,1\right),\qquad
w_i=\ind[r_i>\kappa]+\ind[0\le r_i\le\kappa]\tilde{w}_i,
\]
where \(\mu\) and \(\sigma\) are computed over all valid progress deltas in the training split. Negative-progress chunks receive zero weight, missing deltas use the fallback weight \(1.0\), and each minibatch is renormalized so the weights sum to the batch size. Because RA-BC does not retrieve dataset-supported remaining steps from a structural reachability graph, its continuous reward signal lacks topological granularity, making it prone to missing local execution errors such as slight misalignments or transient intra-chunk regressions.

\paragraph{AWR.}
We implement AWR following the ALOE-style policy in our LeRobot codebase. The baseline has two stages. First, the actor is initialized from the same supervised VLA checkpoint and frozen while training an offline action-chunk critic. The critic is a Transformer over state, image tokens, and the full action chunk, with hidden size 256, 6 layers, 8 attention heads, and two Q readout tokens; we take the minimum over the two heads as the pessimistic value \(Q_{\mathrm{pess}}\). For a chunk with valid horizon \(h\), the training reward is \(-1\) per executed frame, with terminal reward \(0\) for successful episodes and \(-L_{\mathrm{fail}}\) for failures, where \(L_{\mathrm{fail}}\) defaults to the mean episode length for the task. The TD target is
\[
y_t
=
\sum_{j=0}^{h-1}\gamma^j r_{t+j}
+
\gamma^h
\min_k Q_{\bar{\theta},k}(s_{t+h},\hat{a}_{t+h:t+h+N-1},g),
\]
with \(\gamma=0.99\); the bootstrap term is removed when the chunk reaches a terminal state. The next action chunk \(\hat{a}\) is sampled from the frozen current policy and cached during replay construction. We optimize the critic for 50k steps with batch size 128, Adam lr \(3\times10^{-4}\), cosine scheduling with 100 warmup steps, gradient clipping at 1.0, and target-critic Polyak update weight \(0.005\). Second, we freeze the critic, score the aggregated policy-training chunks, and run a weighted behavior-cloning update from the same BC initialization. The critic defines the chunk advantage
\[
A^\pi = Q_{\mathrm{pess}}(s_t,a_{t:t+N-1},g)-V^\pi(s_t,g),
\]
where $V^\pi$ is estimated by evaluating action chunks sampled from the current policy with the same critic. We convert the advantage into a clipped exponential weight,
\[
w_t^{\mathrm{AWR}}=\exp(\mathrm{clip}(A^\pi/\beta_{\mathrm{AWR}},-\epsilon_{\mathrm{clip}},\epsilon_{\mathrm{clip}})),
\]
batch-normalize the weights to mean one, and use them to reweight the standard behavior-cloning objective. The temperature \(\beta_{\mathrm{AWR}}\) and clipping bound \(\epsilon_{\mathrm{clip}}\) are fixed globally for all tasks and rounds rather than tuned per task.

\section{Graph Diagnostics, Representation Ablations, and Hyperparameter Analysis}
\label{app:D}
This section explains why the graph works, why MCR is used for node representation, how robust the method is to graph and weighting hyperparameters, and how shortest-path visualizations support the consequence-based weighting claim.

\subsection{Hyperparameter Analysis}
\label{app:hyperparameter_analysis}

GR-BC contains three groups of hyperparameters: graph construction, progress normalization, and chunk-level weighting. This appendix reports diagnostic sensitivity checks on graph connectivity and chunk weights, followed by the default ranges used in our experiments. These checks are graph-level and weight-level diagnostics on the Assembly Model rollout set; they are not an exhaustive policy-retraining sweep over every hyperparameter.

\begin{table}[t]
\centering
\caption{Hyperparameters and recommended ranges for GR-BC.}
\label{tab:hyperparameter_analysis}
\scriptsize
\setlength{\tabcolsep}{2.2pt}
\renewcommand{\arraystretch}{1.08}
\begin{tabular}{@{}lp{0.25\linewidth}p{0.18\linewidth}p{0.39\linewidth}@{}}
\toprule
Parameter & Meaning & Recommended range & Effect \\
\midrule
\(k\) &
Temporal subsampling interval &
10--20 frames &
Controls graph density. Smaller values preserve fine contact dynamics but increase graph size; larger values reduce cost but may miss short regressions. \\

\(K\) &
Mutual KNN size per camera view &
10--30 &
Controls similarity-edge recall. Larger values improve connectivity but may introduce false visual neighbors. \\

\(\alpha_{\mathrm{edge}}\) &
Adaptive similarity-edge threshold &
0.8--1.2 &
Controls cross-trajectory stitching. Larger values improve failure reachability; smaller values give more conservative edges. \\

\(\epsilon_c\) &
Stability constant for adaptive length &
\(10^{-5}\)--\(10^{-3}\) &
Prevents unstable normalization when local visual scale is very small. \\

\(R_{\max}\) &
Normalization scale for remaining steps &
95th percentile; 95--99th percentile recommended range &
Maps remaining steps to normalized progress. A small value saturates progress early; a large value compresses progress differences. \\

\(N\) &
Action chunk horizon &
20--40 frames &
Defines the lookahead for chunk consequence. Smaller values give local credit; larger values capture longer effects but may mix behaviors. \\

\(\delta_{\min}\) &
Minimum net progress threshold &
0.001--0.005 &
Filters stagnant chunks caused by interpolation or visual noise. \\

\(\eta\) &
Maximum allowed progress regression &
0.02--0.05 &
Controls tolerance to transient regressions. Smaller values suppress unstable chunks; larger values keep more recovery behavior. \\

\(\beta_w\) &
Progress bonus amplitude &
0.3--0.8 &
Controls the strength of reweighting among accepted chunks. \\

\(\gamma\) &
Progress sensitivity &
10--30 &
Controls how sharply weights respond to small progress changes. \\

\(w_{\min}\) &
Lower accepted-branch weight &
0.3--0.7 &
Prevents accepted slow-progress chunks from being ignored. \\

\(w_{\max}\) &
Upper accepted-branch weight &
1.3--2.0 &
Prevents a few high-progress chunks from dominating training. \\
\bottomrule
\end{tabular}
\end{table}

Table~\ref{tab:hyperparameter_sensitivity} reports two compact diagnostics supporting the ranges in Table~\ref{tab:hyperparameter_analysis}. The left panel varies the temporal subsampling interval \(k\) under the same gripper-filtered graph rule. Smaller \(k\) increases graph size and edge count; \(k=10\)--15 keeps high failure-rollout reachability, and \(k=15\) is our default cost--coverage tradeoff. The right panel varies the chunk-weight rule, where \(w=0\) if \(\Delta p_{30}\le\delta_{\min}\), otherwise \(w=\mathrm{clip}(1+\beta_w\tanh(\gamma\Delta p_{30}),0.5,1.5)\). Increasing \(\delta_{\min}\) makes the accepted set more selective and raises the zero-weight ratio, while larger \(\beta_w\) and \(\gamma\) sharpen weight separation but remain bounded by \(w_{\max}=1.5\).

\begin{table}[t]
\centering
\caption{Hyperparameter sensitivity diagnostics.}
\label{tab:hyperparameter_sensitivity}
\label{tab:subsampling_sensitivity}
\label{tab:chunk_weight_sensitivity}
\scriptsize
\renewcommand{\arraystretch}{1.05}
\begin{minipage}[t]{0.42\linewidth}
\centering
\vspace{0pt}
\textbf{Temporal subsampling}
\vspace{2pt}
\resizebox{\linewidth}{!}{
\begin{tabular}{@{}rrrrrr@{}}
\toprule
\(k\) & Sim. edges & Fail reach. & \(R_{\max}\) & Zero wt. & High wt. \\
\midrule
5 & 62,349 & 95.8\% & 125.0 & 26.8\% & 1.9\% \\
10 & 30,221 & 97.5\% & 194.0 & 26.4\% & 21.7\% \\
15 & 20,229 & 98.3\% & 255.0 & 24.6\% & 38.3\% \\
\bottomrule
\end{tabular}}
\end{minipage}
\hfill
\begin{minipage}[t]{0.55\linewidth}
\centering
\vspace{0pt}
\textbf{Chunk weighting}
\vspace{2pt}
\resizebox{\linewidth}{!}{
\begin{tabular}{@{}rrrrrrrr@{}}
\toprule
\(\delta_{\min}\) & \(\beta_w\) & \(\gamma\) & Accepted & Zero wt. & Mean wt. & \(>1\) wt. & p90 / max \\
\midrule
0.003 & 0.3 & 10 & 72.7\% & 27.3\% & 0.811 & 72.7\% & 1.195 / 1.228 \\
0.005 & 0.5 & 20 & 69.6\% & 30.4\% & 0.915 & 69.6\% & 1.457 / 1.482 \\
0.010 & 0.5 & 20 & 63.2\% & 36.8\% & 0.845 & 63.2\% & 1.457 / 1.482 \\
\bottomrule
\end{tabular}}
\end{minipage}
\end{table}

\subsection{Encoder Comparison: R3M, MCR, and V-JEPA2.1}
\label{sec:encoder_comparison}

Graph construction depends on whether the representation groups physically compatible manipulation states together. We compare MCR~\citep{jiang2025robots} with R3M~\citep{nair2022r3m} and two V-JEPA2.1 visual encoders~\citep{mur2026v}. We focus on graph-level diagnostics, since visually plausible retrieval can still introduce harmful shortcuts when the retrieved states are dynamically incompatible. Figure~\ref{fig:encoder_compare} provides a qualitative overview of the resulting graph structure and progress behavior.

\begin{figure}[ht]
  \centering
  \includegraphics[width=0.98\linewidth]{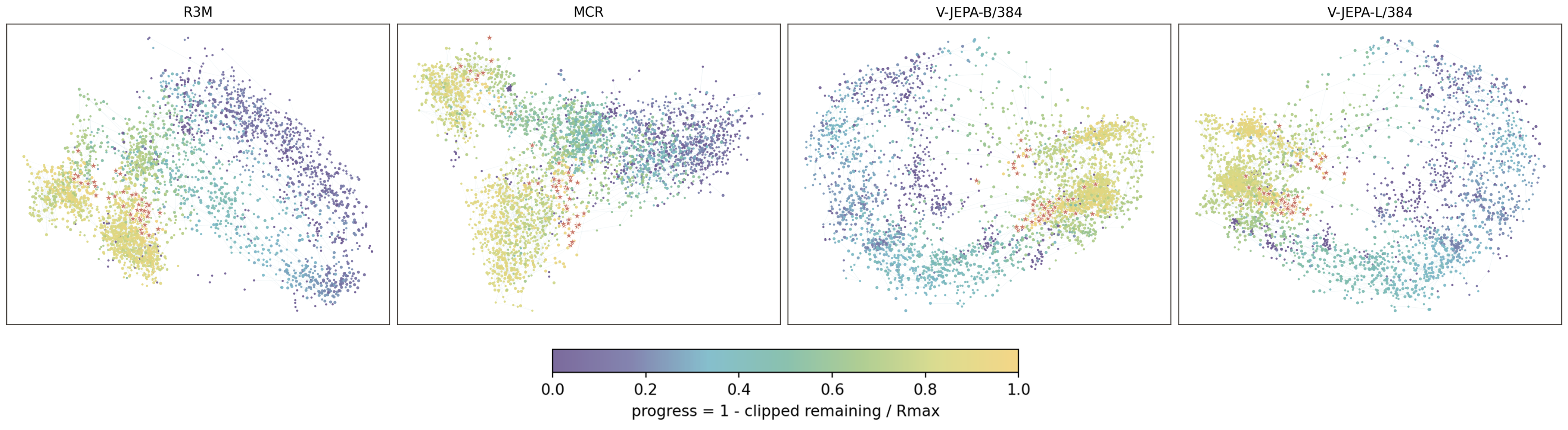}
  \caption{Encoder comparison overview on the Assembly Model rollout dataset. The visualization compares graph structure and progress behavior across R3M, MCR, and V-JEPA2.1 encoders under the same graph-construction rule.}
  \label{fig:encoder_compare}
\end{figure}

\begin{table}[ht]
\centering
\caption{
Encoder comparison on the Assembly Model rollout dataset. All methods use stride $15$, history length $0$, and the same multi-view consensus and gripper-consistency check. Arrows indicate the preferred direction.
}
\label{tab:encoder_comparison}
\resizebox{\linewidth}{!}{
\begin{tabular}{lrrrrrrr}
\toprule
\textbf{Encoder} &
\textbf{Sim. edges $\downarrow$} &
\textbf{$R_{\max}$} &
\textbf{Reachable $\uparrow$} &
\textbf{Mean regression $\downarrow$} &
\textbf{Total variation $\downarrow$} &
\textbf{End progress $\uparrow$} &
\textbf{Encode time $\downarrow$} \\
\midrule
R3M & 12913 & 145.1 & 96.3\% & 0.388 & 1.420 & 0.745 & 60.2s \\
\rowcolor{gray!15}
MCR & 9973 & 223.5 & 94.4\% & 0.437 & 1.289 & 0.713 & 64.3s \\
V-JEPA2.1-B/384 & 19310 & 129.4 & 96.9\% & 0.656 & 1.688 & 0.567 & 123.7s \\
V-JEPA2.1-L/384 & 18980 & 131.8 & 95.1\% & 0.642 & 1.655 & 0.562 & 176.9s \\
\bottomrule
\end{tabular}
}
\end{table}

Table~\ref{tab:encoder_comparison} summarizes the graph diagnostics. ``Sim. edges'' counts accepted non-temporal similarity edges after multi-view consensus and gripper-consistency checks. Unless otherwise specified, \(R_{\max}\) is the 95th percentile of finite shortest-path distances to success and is used for progress normalization; we recommend 95--99\% as a robust range. ``Reachable'' measures the fraction of nodes connected to a success node. ``Mean regression'' and ``Total variation'' quantify backward progress jumps and overall progress smoothness. ``End progress'' is the average final episode progress, and ``Encode time'' is the feature extraction time for all sampled nodes and three camera views.

R3M is competitive and even better on mean regression and encoding speed, while MCR yields a more conservative graph with lower total variation and fewer accepted similarity shortcuts. Both V-JEPA2.1 encoders are substantially slower and produce denser graphs with larger regression and variation. We choose MCR because it gives the most conservative graph among the compared encoders, with the lowest total variation and acceptable encoding cost.

\subsection{Multiple Successful Terminal Clusters}
\label{sec:multi_success_sinks}

We use Assembly Model as a diagnostic for multi-terminal success. The task admits multiple correct terminal configurations, and our analysis recovers this structure directly: successful terminal nodes form two visually and physically distinct groups, as shown by the feature-space visualization and representative camera frames in Figure~\ref{fig:multi_success_sink_graph}. Our graph connects all successful terminals to the virtual sink \(g_\star\), so \(\Rrec(s)\) is the shortest cost to the nearest reachable success mode rather than to a single hand-specified goal. This supports the robustness of our pipeline under multi-modal success definitions.

The value \(p\in[0,1]\) indicates the normalized trajectory progress. 
Labels \texttt{m0} and \texttt{m1} correspond to two distinct successful execution modes, while \texttt{fail} denotes states from failed trajectories. 

\subsection{Shortest-Path Visualization}
\label{sec:shortest_path_visualization}

\begin{wrapfigure}{r}{0.48\linewidth}
  \centering
  \includegraphics[width=\linewidth]{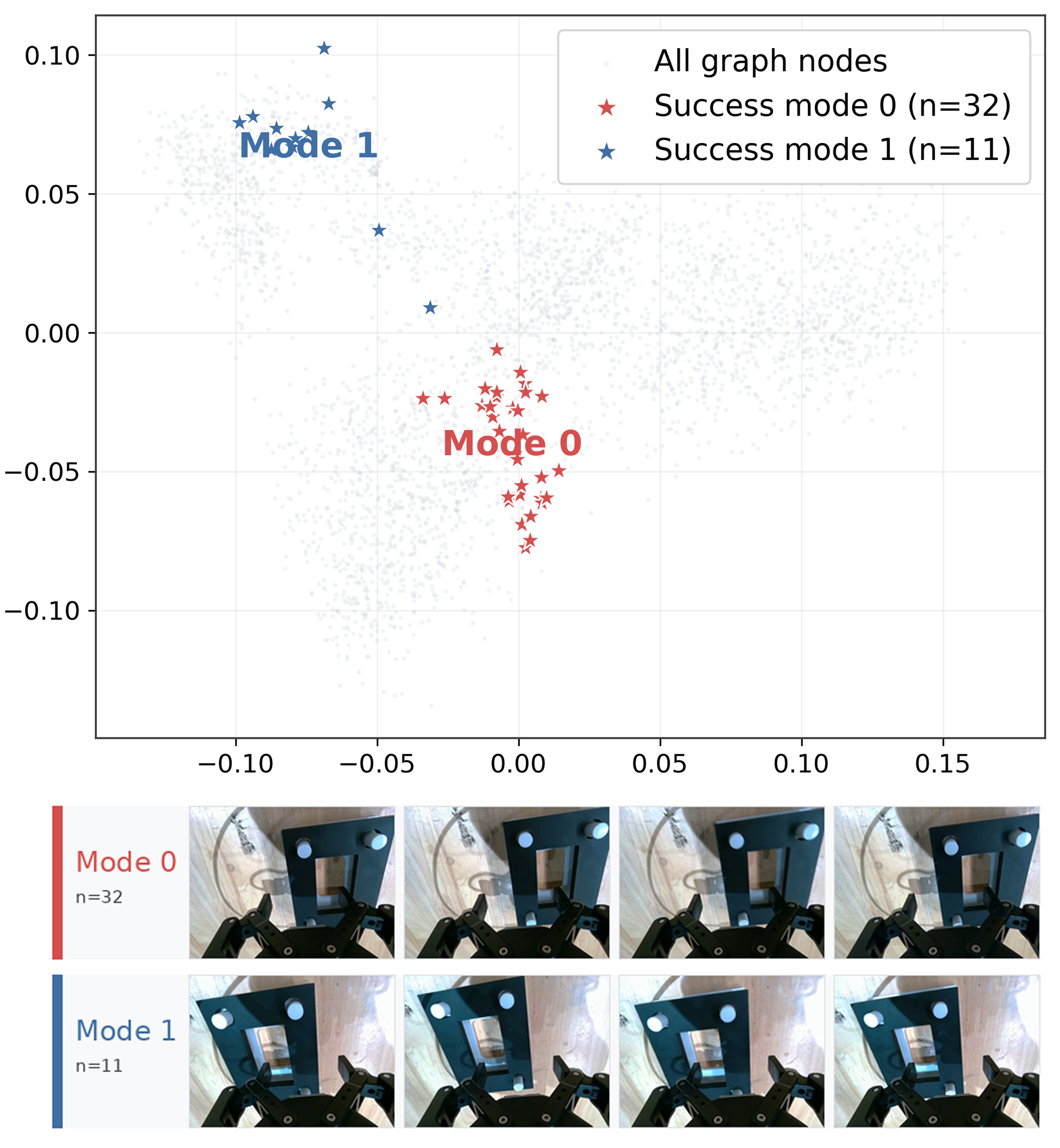}
  \caption{Multiple Success Modes on Assembly Model.}
  \label{fig:multi_success_sink_graph}
  \vspace{-6pt}
\end{wrapfigure}

Figure~\ref{fig:shortest_path_visualization} visualizes one recovered path from a failed state to a successful terminal state. In the visualization, \(E\) denotes the episode index and \(F\) denotes the frame index within that episode. 
The edge label \(T\) denotes a temporal transition within an episode, and \(S\) denotes a similarity-based bridge across episodes. 
\texttt{Start} and \texttt{Goal} mark the queried state and the recovered successful terminal state, respectively.

This example shows that our pipeline does not discard failed data. 
Instead, failed states can be connected through similarity bridges to useful recovery segments and then routed toward successful terminals. 
Moreover, the path crosses different episodes and success modes, indicating that the method can exploit shared sub-trajectories where different successful strategies intersect. 
As a result, the graph-based weighting captures reusable structure from both successful and failed demonstrations, enabling effective consequence-based credit assignment.

\begin{figure}[h]
  \centering
  \includegraphics[width=0.98\linewidth]{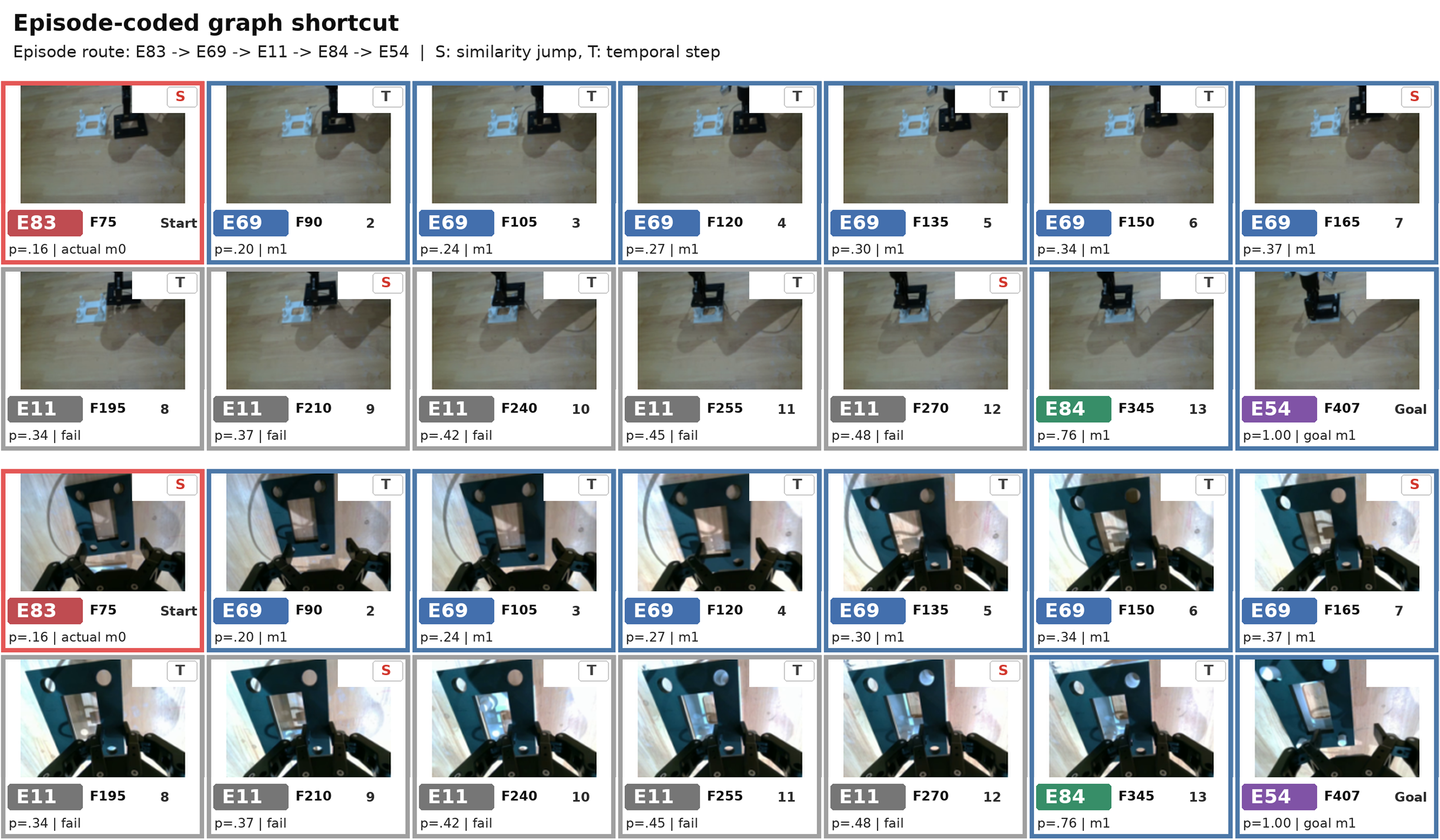}
  \caption{Shortest-path visualization from a failed state to a successful terminal state. The path alternates between temporal transitions and similarity bridges, showing how failed data can be routed through reusable recovery segments.}
  \label{fig:shortest_path_visualization}
\end{figure}

\section{Additional Progress Visualization}

\begin{figure}[h]
  \centering
  \includegraphics[width=0.98\linewidth]{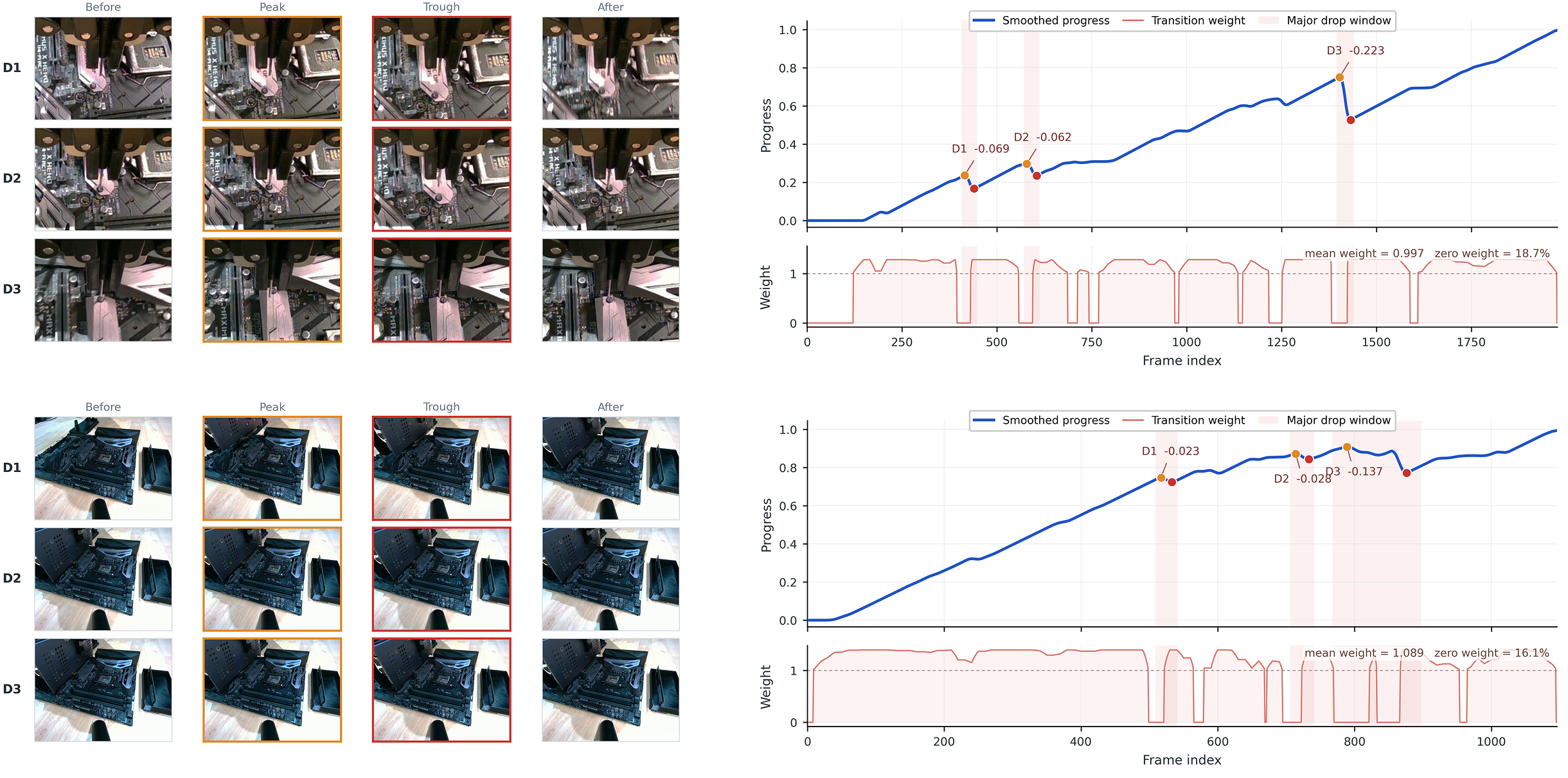}
  \caption{Examples of Progress prediction on demonstration data.}
  \label{fig:progress_demo_examples}
\end{figure}

\begin{figure}[h]
  \centering
  \includegraphics[width=0.98\linewidth]{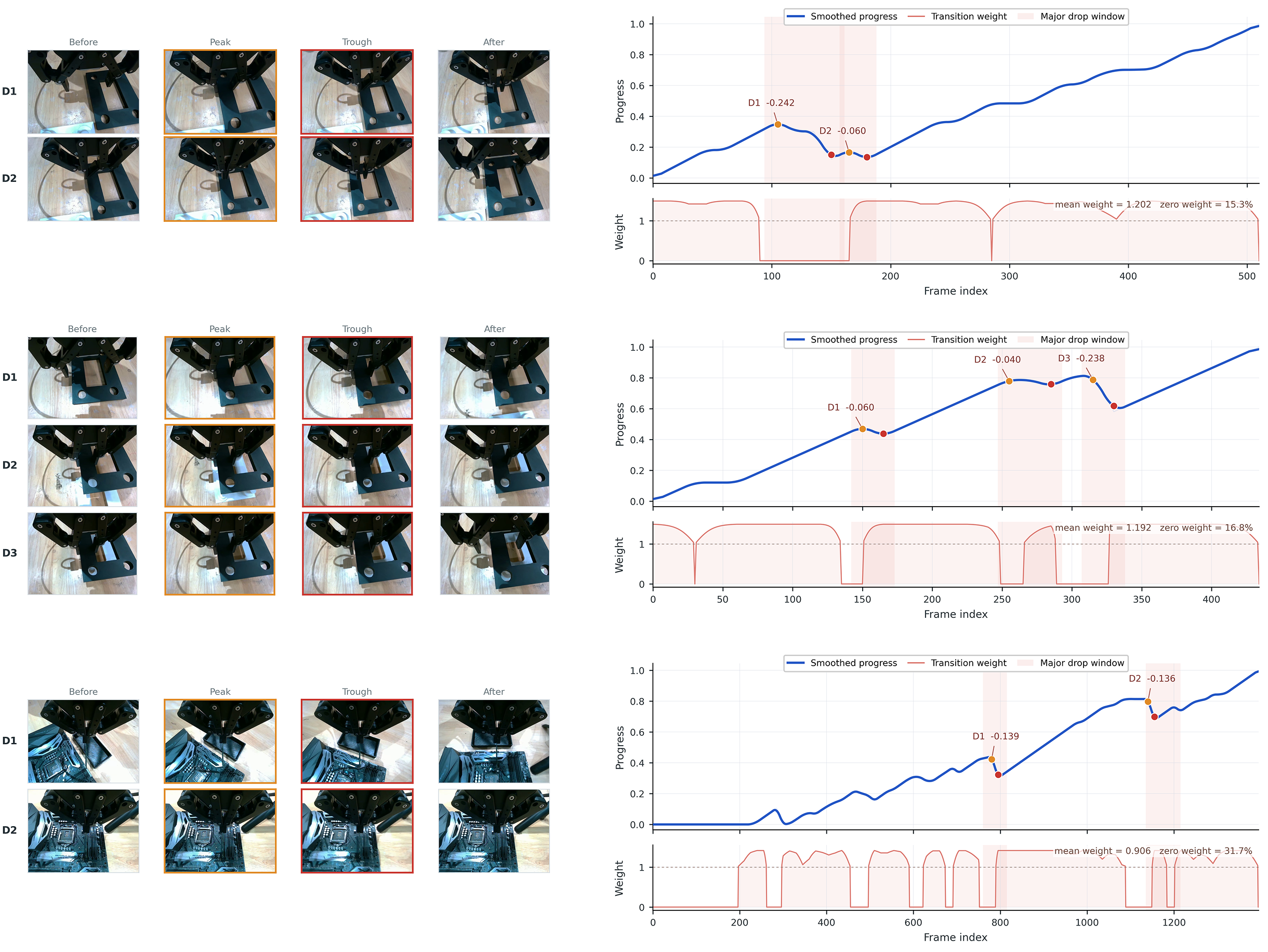}
  \caption{Examples of Progress prediction on policy rollouts.}
  \label{fig:progress_rollout_examples}
\end{figure}

\clearpage
\endgroup

\end{document}